\pdfoutput=1
\documentclass[11pt]{article}
\usepackage[final]{acl}

\usepackage{times}
\usepackage{latexsym}

\usepackage[T1]{fontenc}

\usepackage[utf8]{inputenc}

\usepackage{microtype}

\usepackage{inconsolata}

\usepackage{graphicx}
\usepackage{booktabs}
\usepackage{array} 
\usepackage{multirow}
\usepackage{subcaption}
\usepackage{arydshln}
\usepackage{makecell}
\usepackage{amsmath}
\usepackage{tcolorbox}

\definecolor{todo}{rgb}{1,0,0}

\definecolor{ik}{rgb}{0,0,1}

\definecolor{care}{rgb}{0.961, 0.412, 0.259}

\usepackage{enumitem}

\title{\textsc{ABCD-LINK}:\\Annotation Bootstrapping for Cross-Document Fine-Grained Links}

\author{
Serwar Basch$^{\textbf{\texttt{1}}}$, Ilia Kuznetsov$^{\textbf{\texttt{1}}}$, Tom Hope$^{\textbf{\texttt{2}}}$, Iryna Gurevych$^{\textbf{\texttt{1}}}$
\\
        \textsuperscript{\textbf{\texttt{1}}}Ubiquitous Knowledge Processing Lab (UKP Lab), \\ Department of Computer Science 
and Hessian Center for AI (hessian.AI), TU Darmstadt \\ 
\textsuperscript{\textbf{\texttt{2}}} Hebrew University of Jerusalem and The Allen Institute for AI (AI2)\\
\href{https://www.ukp.tu-darmstadt.de}{www.ukp.tu-darmstadt.de}
}

\begin{document}
\maketitle
\begin{abstract}
Understanding fine-grained links between documents is crucial for many applications, yet progress is limited by the lack of efficient methods for data curation. To address this limitation, we introduce a domain-agnostic framework for bootstrapping sentence-level cross-document links from scratch. Our approach (1) generates and validates semi-synthetic datasets of linked documents, (2) uses these datasets to benchmark and shortlist the best-performing linking approaches, and (3) applies the shortlisted methods in large-scale human-in-the-loop annotation of natural text pairs. We apply the framework in two distinct domains – peer review and news – and show that combining retrieval models with LLMs achieves a 73\% human approval rate for suggested links, more than doubling the acceptance of strong retrievers alone. Our framework allows users to produce novel datasets that enable systematic study of cross-document understanding, supporting downstream tasks such as media framing analysis and peer review assessment. All code, data, and annotation protocols are released to facilitate future research.\footnote{\href{https://github.com/UKPLab/eacl2026-abcd-link}{https://github.com/UKPLab/eacl2026-abcd-link}}
\end{abstract}

\begin{figure}[!t]
\centering
  \includegraphics[width=0.8\linewidth]{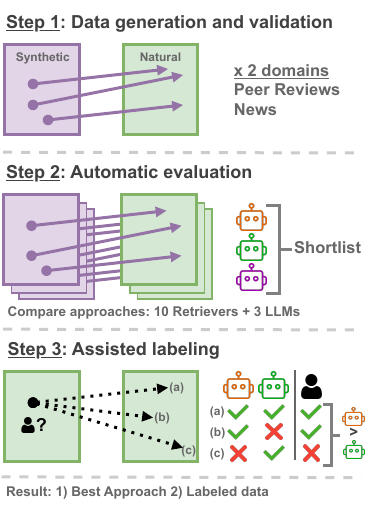}
  \caption{Framework overview.}
  \label{fig:framework}
\end{figure}

\section{Introduction}

Documents rarely exist in isolation. In many practical scenarios, understanding one document requires reasoning over its relationships with others. Fact-checkers might trace a claim to the specific sentences across multiple articles that support or contradict it \citep{thorne-etal-2018-fever, wadden-etal-2020-fact, chen-etal-2024-metasumperceiver}. Authors and meta-reviewers reading peer reviews need to connect reviewer comments to the paper to decide on the course of action \citep{kuznetsov-etal-2022-revise, darcy-etal-2024-aries}. News analysts might use paraphrased or ideologically reframed sentences across different outlets to gain insights into reporting bias and source alignment \citep{giorgi-etal-2023-open}. Yet finding such fine-grained relations can be cognitively demanding, motivating the need for machine assistance. While tasks like cross-document coreference resolution offer some support, cross-document relations are much more diverse, necessitating the development of general-purpose approaches to cross-document analysis.

Following \citet{kuznetsov-etal-2022-revise}, we call fine-grained relations between documents \textit{links}, and the general task of identifying them \textit{linking}. While automatic linking holds great potential, the progress has been limited due to the manual effort required to annotate and evaluate links, and the diversity of link types across domains. Thus, labeled corpora of links are scarce, preventing the development and evaluation of automated linking approaches at scale.

To move beyond the bottleneck of full manual annotation, we propose a general framework for bootstrapping sentence-level cross-document linking without the need for pre-existing labeled data (Figure~\ref{fig:framework}). While links might be hard for humans to \textit{detect}, linked documents are easier to \textit{generate} with state of the art LLMs. Building upon this insight, we (1) propose a method to generate and validate semi-synthetic corpora of linked documents. We use this data to (2) automatically evaluate a wide range of approaches to arrive at a shortlist of best-performing linking approaches. Finally, we use the best-performing approaches to (3) assist humans in labeling natural, non-synthetic document pairs in a large-scale human-in-the-loop evaluation study. Our framework allows practitioners to efficiently select the best linking approach given a domain and link type, while producing manually labeled many-to-many cross-document linking datasets for downstream applications.

We assess our framework by applying it in two distinct domains, peer reviews and news articles, each with its own task-specific link type and underlying data. In total, we contribute:
\setlist{nolistsep}
\begin{itemize}[noitemsep]
    \item A novel framework for efficient human-in-the-loop evaluation and labeling of links;
    \item A data generation approach for creating semi-synthetic datasets of linked documents in the domain of interest, incl. human validation;
    \item A large-scale analysis of state-of-the-art linking approaches, incl. a novel human evaluation protocol;
    \item Two new manually annotated non-synthetic corpora of cross-document links in peer review and news.
\end{itemize}

We release the code, datasets, and annotation protocols to support future research on sentence-level cross-document linking and its applications.

\section{Related Work}

Numerous NLP tasks study or rely on \textbf{cross-document relations}, including cross-document coreference resolution (CDCR) \citep{DBLP:conf/ijcnlp/MiyabeTO08, cybulska-vossen-2014-using, ravenscroft-etal-2021-cd}, fact-checking \citep{thorne-etal-2018-fever, wadden-etal-2020-fact, chen-etal-2024-metasumperceiver}, scholarly peer review analysis \citep{kuznetsov-etal-2022-revise, darcy-etal-2024-aries}, citation/source detection \citep{syed-etal-2023-citance, liang-etal-2024-fine}, and cross-document question answering \cite{lin-etal-2025-mebench}. While prior work focuses on limited relation types and relies on manual labeling, we contribute a general, structured framework for bootstrapping linking from scratch in new domains and for new link types. We evaluate our framework in two domains: peer reviews and news, which target challenging and underrepresented forms of linking, such as subjective evaluation and ideological framing, as opposed to encyclopedic and factual relations found in Wikipedia \citep{feith-etal-2024-entity}. Crucially, unlike CDCR where the main goal is to resolve and cluster entity and event mentions, linking focuses on detecting a broad spectrum of sentence-level relationships, such as paraphrases, quotes and implicit references.

From a \textbf{machine-assisted annotation} perspective, prior work reduces labeling effort by pre-selecting candidate links with heuristics \cite{ravenscroft-etal-2021-cd}, or simple retrieval methods, e.g.\ cosine similarity with a fixed cutoff \citep{kuznetsov-etal-2022-revise}. Our framework generalizes from this idea: we automatically benchmark various zero-shot approaches, and validate the results through human evaluation to select the best approach for the task and domain at hand. In that, our work contributes to the study of \textbf{synthetic data} in NLP \citep{ding-etal-2024-data, liu-etal-2022-wanli, zhao-etal-2025-beyond, josifoski-etal-2023-exploiting, veselovsky2023generating}.
\citet{moller-etal-2024-parrot} and \citet{kazemi-etal-2025-synthetic} find that synthetic data can match or complement human-labeled corpora in low-resource settings. 
To the best of our knowledge, we are the first to use synthetic data to study cross-document links. The use of synthetic data carries risks \citep{pmlr-v202-van-breugel23a, DBLP:journals/corr/abs-2302-04062}. To mitigate them, our framework only uses synthetic data to automatically evaluate different linking approaches, employs a human validation step, and supplements evaluations on synthetic data with focused human evaluation, contributing to best practices in the use of synthetic data in NLP.

Human judgment is essential for evaluating NLP systems, yet difficult to elicit reliably. There is growing interest in structured \textbf{human evaluation protocols}, particularly for tasks involving model-assisted annotation or subjective decision-making. Prior work examined inter-annotator calibration over time \citep{uma-etal-2021-semeval}, preference-based assessment of model outputs \citep{clark-etal-2021-thats}, and the influence of interface design on annotation quality \citep{klie-etal-2018-inception}. In the context of LLM evaluation, human-in-the-loop setups have proven essential for capturing fine-grained distinctions and improving label quality \citep{10.5555/3666122.3668522, min-etal-2025-multi}. Here, we contribute a novel and efficient annotation protocol that combines candidate pre-selection with a ranking-based comparison for a reliable assessment of top-performing approaches.

\section{Setup}

We focus on \emph{sentence-level linking} cast as a sentence pair classification task, representing a trade-off between task complexity and utility: a more narrow unit of analysis restricts the available context, while the links between broader units can be reconstructed by aggregating consecutive sentences.\footnote{While broader argumentative or narrative relations that span multiple consecutive sentences do exist, in practice, many-to-many relations naturally emerge as clusters of consecutive sentence-level links. We provide an exploratory analysis of such clusters in Appendix~\ref{app:group_links}} Given two documents, the source document \( A = \{a_1, \ldots, a_N\} \) and the target document \( B = \{b_1, \ldots, b_M\} \), the goal is to predict for each cross-document sentence pair \( (a_i, b_j) \) whether a relation \( r \) holds between them. The definition accommodates many-to-many links, both directed (e.g. a review sentence criticizing a claim in a paper) and undirected (e.g. sentences in two new articles addressing the same fact). Linking can be seen as a special case of information retrieval, where $a_i$ serves as a query to retrieve relevant sentences from $B$. However, linking is constrained by a task-specific relationship holding between sentences, and might require documents $A$ and $B$ for contextualizing the linking decisions.

There are many potential approaches to perform automatic linking in a zero-shot fashion, from simple relation-agnostic cosine similarity to classification with LLMs. Yet, labeled datasets of links are scarce, and \textit{evaluating and comparing} the performance of different linking approaches on a wide range of link types and domains via human evaluation is infeasible. Our framework addresses this limitation by generating and validating semi-synthetic data (Section \ref{sec:data}) to arrive at a shortlist of best-performing linking approaches (Section \ref{sec:sys_eval}) which are then used in a focused human evaluation and annotation study (Section \ref{sec:human_eval}). 

The framework is designed to be domain-agnostic in that it does not depend on any domain-specific characteristics like writing style, relationship type or the presence of explicit references such as citations. To demonstrate this, we apply the framework in two distinct practical scenarios: in \texttt{REVIEWS}, we link peer reviews to their papers, and in \texttt{NEWS}, we link pairs of news articles. While the framework itself aims to be domain-agnostic, the particular link types (and cutoff values, see Section~\ref{sec:sys_eval}) are flexible and defined per domain. In \texttt{REVIEWS}, we define a link as a sentence in a review that comments on, critiques, or supports a specific sentence in the corresponding paper. In \texttt{NEWS}, a link connects two sentences from different articles that convey the same or closely related factual content, typically reflect semantic equivalence or paraphrastic overlap, even if the framing or tone differs between the two sources.

We showcase our framework on two distinct practical scenarios: in \texttt{REVIEWS}, we link peer reviews to their papers, and in \texttt{NEWS}, we link pairs of news articles. In \texttt{REVIEWS}, we define a link as a sentence in a review that comments on, critiques, or supports a specific sentence in the corresponding paper. In \texttt{NEWS}, a link connects two sentences from different articles that convey the same or closely related factual content, typically reflect semantic equivalence or paraphrastic overlap, even if the framing or tone differs between the two sources.

\section{Synthetic data}
\label{sec:data}

\subsection{Generation}
We hypothesize that while links in existing document pairs can be time-consuming to annotate, they are relatively easy to generate. Based on this intuition, we construct a semi-synthetic linking dataset for each of our target application domains using a non-synthetic, natural document as target, and prompting an LLM to generate a synthetic source document that links to the target on the sentence level. For this experiment, we use DeepSeek-R1 \citep{DBLP:journals/corr/abs-2501-12948} due to its strong instruction-following performance and ability to handle long context. To avoid potential bias \cite{liu-etal-2024-llms-narcissistic}, we explicitly exclude DeepSeek-R1 from downstream experiments, and solely use it for synthetic data generation.
While strictly speaking only the source documents in our data are synthetic, for simplicity, we further refer to the resulting data as our \textit{synthetic data}.

\begin{table}[t]
\small
\centering
\begin{tabular}{l@{\hskip 6pt}c@{\hskip 6pt}c@{\hskip 6pt}c@{\hskip 6pt}c}
\toprule
\textbf{\texttt{Stat}} &
\makecell{\cellcolor{orange!30}\texttt{\textbf{NEWS}} \\ \texttt{\textbf{ECB+}}} &
\makecell{\cellcolor{orange!30}\texttt{\textbf{NEWS}} \\ \cellcolor{violet!30}\texttt{\textbf{SYNTH}}} &
\makecell{\cellcolor{gray!20!cyan!30}\texttt{\textbf{REVIEWS}} \\ \cellcolor{violet!30}\texttt{\textbf{SYNTH}}} &
\makecell{\cellcolor{gray!20!cyan!30}\texttt{\textbf{REVIEWS}} \\ \texttt{\textbf{F1000}}} \\
\midrule
\texttt{Doc Pairs} & \texttt{2505} & \texttt{346} & \texttt{200} & \texttt{211} \\
\texttt{Number of Links} & \texttt{9383} & \texttt{2323} & \texttt{2181} & \texttt{1205} \\
\\[-1.6ex]
\cdashline{1-5}
\\[-1ex]
\texttt{Avg. Sents (Src)} & \texttt{17.7} & \texttt{12.0} & \texttt{10.5} & \texttt{22.4} \\
\texttt{Avg. Sents (Tgt)} & \texttt{18.2} & \texttt{13.7} & \texttt{130.8} & \texttt{145.0} \\
\\[-1.6ex]
\cdashline{1-5}
\\[-1ex]
\texttt{Avg. Links (Src)} & \texttt{2.06} & \texttt{4.97} & \texttt{5.14} & \texttt{4.84} \\
\texttt{Avg. Links (Tgt)} & \texttt{3.75} & \texttt{6.71} & \texttt{10.9} & \texttt{5.71} \\
\\[-1.6ex]
\cdashline{1-5}
\\[-1ex]
\texttt{Tgt/Src Ratio} & \texttt{1.82} & \texttt{1.17} & \texttt{2.14} & \texttt{1.18} \\
\texttt{Src/Tgt Ratio} & \texttt{1.86} & \texttt{1.07} & \texttt{1.03} & \texttt{1.15} \\
\bottomrule
\end{tabular}
\caption{Dataset statistics: number of document pairs, number of links, average number of sentences in source and target documents, average number of linked sentences in source and target document, and link density.}
\label{tab:dataset_stats}
\end{table}

\textbf{\texttt{REVIEWS-SYNTH}} builds upon the NLPeer dataset \cite{dycke-etal-2023-nlpeer}, which includes EMNLP24 papers pre-segmented into sentences. To mitigate long-context issues \citep{levy-etal-2024-task} and have a comparable size to \texttt{REVIEW-F1000}, we select the 200 shortest papers by sentence count and prompt the model to generate a peer review for each. 
The prompt instructs the LLM to write review sentences and link them to specific sentences in the paper, simulating naturally occurring reviewer comments.

To match the number of links in \texttt{REVIEWS-SYNTH}, for \textbf{\texttt{NEWS-SYNTH}} we sample 346 news articles from the WikinewsSum dataset \citep{calizzano-etal-2022-generating}. To reduce noise in the documents, we clean each article using GPT-4o-mini to remove scraping artifacts. 
We then segment the text into sentences, and prompt the LLM to produce a related article on the same topic, written with a different tone or editorial perspective, and instructing the model to link sentences in the generated article to sentences in the original article, resulting in cross-document links. This simulates a use case in media analysis where the same events reported by different outlets are compared to detect bias and framing. See Appendix~\ref{app:synth_data_gen} for more details, prompt templates and example links for both datasets.

To contextualize our synthetic datasets and experimental results, we derive two additional datasets from existing cross-document corpora. \textbf{\texttt{NEWS-ECB+}} is derived from the \texttt{ECB+} corpus \citep{cybulska-vossen-2014-using} designed for cross-document coreference resolution. We repurpose this dataset for sentence-level linking by aligning sentence pairs across documents that share event mentions in the same cluster.
\textbf{\texttt{REVIEWS-F1000}} leverages the \texttt{F1000RD} dataset introduced by \citet{kuznetsov-etal-2022-revise}, which links peer review sentences to their corresponding paper sentences. Although its annotation design restricts the choices of possible links, it is the most structurally similar resource to our task. Dataset conversion details are provided in Appendix~\ref{app:data_conversion}. Table~\ref{tab:dataset_stats} provides key statistics on the synthetic and converted datasets.

\begin{figure}[!t]
  \includegraphics[width=\linewidth]{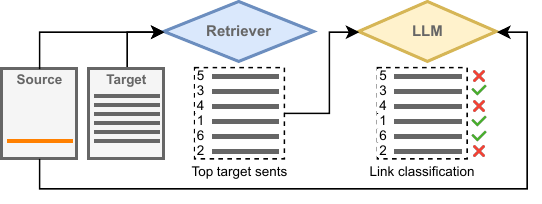}
  \caption {\texttt{R+LLM} Setup.}
  \label{fig:llm_setup}
\end{figure}

\subsection{Validation}
\label{sec:validation}
We ensure the quality of the synthetic data, and explore potential biases and hallucinations in machine-generated data, through automatic and manual inspection. At the document level, we compare lexical diversity, subjectivity, and complexity (Flesch-Kincaid score~\citep{flesch1948new}). In \texttt{NEWS-SYNTH}, synthetic articles are more lexically diverse, similarly subjective, and slightly more lexically complex than naturally occurring articles. In \texttt{REVIEWS-SYNTH}, synthetic reviews are less diverse but more complex, reflecting LLMs’ preference for concise, formal phrasing. Subjectivity remains comparable across domains. Overall, synthetic documents align well with natural ones in most stylistic dimensions (more details in Appendix~\ref{app:quan_criteria})

\begin{table*}[!t]
\centering
\small
\begin{tabular}{lcc!{\color{gray}\vrule width 0.6pt}cc!{\color{gray}\vrule width 0.6pt}cc!{\color{gray}\vrule width 0.6pt}cc!{\color{gray}\vrule width 0.6pt}c}
\toprule
& \multicolumn{2}{c}{\texttt{\textbf{NEWS-ECB+}}} 
& \multicolumn{2}{c}{\texttt{\textbf{NEWS-SYNTH}}} 
& \multicolumn{2}{c}{\texttt{\textbf{REVIEWS-SYNTH}}} 
& \multicolumn{2}{c}{\texttt{\textbf{REVIEWS-F1000}}} 
&  \texttt{\textbf{Overall}}\\
\cmidrule(lr){2-10}
\texttt{\textbf{Model}} & \texttt{Avg. F1} & \texttt{R@10} & \texttt{Avg. F1} & \texttt{R@10} & \texttt{Avg. F1} & \texttt{R@20} & \texttt{Avg. F1} & \texttt{R@20} & \texttt{Avg. F1}\\
\midrule
\multicolumn{10}{c}{\texttt{\textbf{Sparse Models}}} \\
\midrule
\texttt{BM25}          & \texttt{37.26} & \texttt{89.40} & \texttt{32.58} & \texttt{94.35} & \texttt{18.91} & \texttt{66.39} & \texttt{29.53} & \texttt{93.90} & \texttt{29.57} \\
\texttt{SPLADEv3}        & \texttt{39.83} & \texttt{93.23} & \texttt{33.04} & \texttt{95.45} & \texttt{19.46} & \texttt{69.83} & \texttt{29.10} & \texttt{95.59} & \texttt{30.36} \\
\texttt{BGE-M3-SPARSE}  & \texttt{38.81} & \texttt{92.21} & \texttt{34.52} & \texttt{96.49} & \texttt{19.73} & \texttt{70.08} & \texttt{29.68} & \texttt{95.38} & \texttt{30.68} \\
\midrule
\multicolumn{10}{c}{\texttt{\textbf{Bi-Encoders}}} \\
\midrule
\texttt{SFR}           & \texttt{43.62} & \texttt{98.01} & \texttt{36.28} & \texttt{97.8} & \texttt{19.20} & \texttt{75.93} & \texttt{25.56} & \texttt{92.05} & \texttt{31.17} \\
\texttt{all-mpnet-base}     & \texttt{42.18} & \texttt{96.52} & \texttt{36.73} & \texttt{98.97} & \texttt{18.09} & \texttt{68.69} & \texttt{24.54} & \texttt{89.16} & \texttt{30.38} \\
\texttt{BGE-M3-DENSE}   & \texttt{42.33} & \texttt{96.23} & \texttt{37.32} & \texttt{99.79} & \texttt{21.52} & \texttt{77.14} & \texttt{29.80} & \texttt{96.14} & \texttt{32.74} \\
\texttt{Contriever}    & \texttt{41.17} & \texttt{95.73} & \texttt{33.17} & \texttt{97.31} & \texttt{17.30} & \texttt{66.69} & \texttt{26.14} & \texttt{93.45} & \texttt{29.45} \\
\texttt{Dragon+}  & \texttt{42.42} & \texttt{96.72} & \texttt{36.99} & \texttt{99.17} & \texttt{21.11} & \texttt{73.50} & \texttt{31.51} & \texttt{97.08} & \texttt{\textbf{33.01}} \\
\midrule
\multicolumn{10}{c}{\texttt{\textbf{Cross-Encoders}}} \\
\midrule
\texttt{BGE-M3-Reranker}      & \texttt{41.39} & \texttt{96.74} & \texttt{36.83} & \texttt{100.0} & \texttt{19.52} & \texttt{77.11} & \texttt{27.22} & \texttt{95.74} & \texttt{31.24} \\
\texttt{ms-marco-MiniLM} & \texttt{41.88} & \texttt{96.27} & \texttt{36.71} & \texttt{98.76} & \texttt{21.26} & \texttt{72.97} & \texttt{31.88} & \texttt{96.65} & \texttt{\underline{32.93}} \\
\bottomrule
\end{tabular}
\caption{Performance of retrieval models across synthetic and converted datasets. Metrics are: average F1 score across all cutoffs ($k \in \{1, 3, 5, 7, 10, 20\}$) and recall at a fixed cutoff (R@10 for news, R@20 for reviews). The overall average F1 is computed across all datasets. Recall cutoffs are selected based on domain-specific document lengths to maximize retrieval coverage. Detailed per-dataset and per-cutoff results are provided in Appendix~\ref{app:detailed_results}.}
\label{tab:retireval_results}
\end{table*}

To further assess the quality of the synthetic data, we conduct a human validation study. For \texttt{NEWS-SYNTH}, we randomly selected 30 natural news articles and their corresponding synthetic counterparts. Four annotators rated each article on a five-point Likert scale for fluency, coherence, realism (i.e., could the article be written by a human), and specificity (i.e., does the article contain specific facts). Synthetic articles match fluency/coherence but score lower on realism (as shown in Figure~\ref{fig:synth_news_annos_discrete} in the Appendix) which we attribute to the models limitation in generating realistic sounding news articles. The synthetic data also shows lower specificity, likely due to LLMs producing more general statements compared to natural articles. While these limitations may affect journalistic authenticity, they do not hinder our linking task where coherence is more important, because it provides better context to reason over.

For \texttt{REVIEWS-SYNTH}, we randomly selected 30 synthetic reviews and retrieved natural reviews for the same papers from the NLPeer dataset. Three NLP PhD students rated each review on fluency, coherence, helpfulness (i.e., would this review be helpful for the paper’s authors), and specificity (i.e., does the review address concrete parts of the paper). Synthetic reviews were rated higher in fluency and coherence, and lower in helpfulness and specificity due to a lack of concrete suggestions and feedback. While not perfect, the results align with our goal of generating reviews that are coherent and linkable. See Appendix~\ref{app:human_val} for further details on the setup.

In summary, while LLM-generated linked document pairs are sometimes less specific or natural, they offer a practical, scalable way to create sentence-level links that can be used for automatic evaluation.

\section{Automatic Evaluation}
\label{sec:sys_eval}

\subsection{Approaches}
Using our synthetic data, we benchmark a broad range of retrievers in a zero-shot setup to shortlist strong candidates for human evaluation. We test sparse models (\texttt{BM25} \citep{INR-019}, \texttt{SPLADEv3} \citep{DBLP:journals/corr/abs-2403-06789}, \texttt{BGE-M3} \citep{chen-etal-2024-m3}), bi-encoders (\texttt{all-mpnet-base-v2}\footnote{\href{https://huggingface.co/sentence-transformers/all-mpnet-base-v2}{sentence-transformers/all-mpnet-base-v2}}, \texttt{SFR-Embedding-Mistral} \citep{meng2024sfrembedding}, \texttt{BGE-M3}, \texttt{Contriever} \citep{izacard2022unsupervised}, \texttt{Dragon+} \citep{lin-etal-2023-train}), and cross-encoders (\texttt{ms-marco-MiniLM-L6-v2}\footnote{\href{https://huggingface.co/cross-encoder/ms-marco-MiniLM-L6-v2}{cross-encoder/ms-marco-MiniLM-L6-v2}}, \texttt{BGE-M3}). Each retriever ranks candidate sentences in the target document based on cosine similarity to a sentence from the source document. We convert ranks into binary decisions via a threshold: all sentences that appear within the top-$k$ are treated as links.

\begin{figure*}[!t]
  \includegraphics[width=\linewidth]{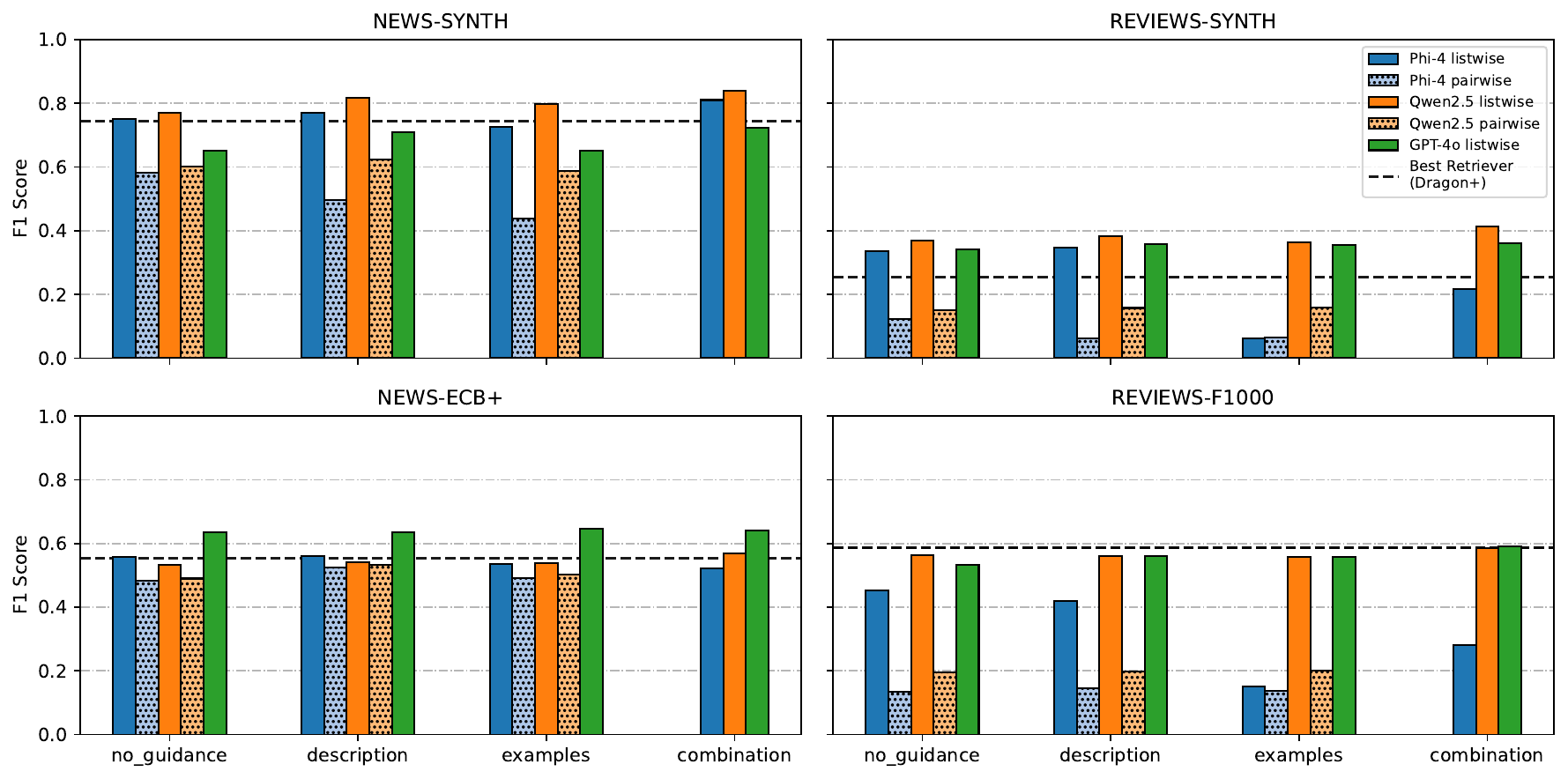}
  \caption {F1 scores across datasets using different prompt configurations and models (Phi-4, Qwen2.5, GPT-4o). Results show a consistent trend across domains with listwise prompting outperforming pairwise prompting. Furthermore, combination prompts (tested listwise only) achieve the highest F1 scores on synthetic datasets, while performance gaps narrow on the converted datasets. These results showcase the interaction between prompt design, model capabilities, and dataset complexity.}
  \label{fig:llm_results}
\end{figure*}

We then extend this setup with LLMs to refine candidate links. The best-performing retriever supplies the top-$k$ target sentences. We choose the $k$ based on the performance of the retrieval models, and the length of the documents, to maximize the ($k=10$ for \texttt{NEWS} domain, $k=20$ for \texttt{REVIEWS} domain). Next, the LLM classifies the source-target pairs as linked or not linked (Figure~\ref{fig:llm_setup}). We use LLMs for their zero-shot ability to adapt to different link types via prompting only, making them well-suited for domains where link semantics vary and labeled data for fine-tuning is unavailable. We further refer to this approach as \texttt{R+LLM} for brevity.

We investigate two prompting setups. \textbf{Pairwise}, where each source--target pair is judged independently, and \textbf{Listwise}, where all $k$ candidates are considered jointly in a single pass. In each setup, we prompt the LLM with both whole documents in four configurations: with a link description, with in-context examples, with both, and with no guidance. For evaluation, we use two open-source models, \texttt{Phi-4 (14B)} \cite{DBLP:journals/corr/abs-2412-08905} and \texttt{Qwen2.5 (32B)} \cite{DBLP:journals/corr/abs-2412-15115}, selected for their SOTA performance within their respective size classes. We also include \texttt{GPT-4o}\footnote{Version gpt-4o-2024-08-06}, as the SOTA closed-source non-reasoning model at the time of writing, to serve as a high-performance reference point. Prompt templates are given in Appendix~\ref{app:llm_prompts}.

\subsection{Results}
\paragraph{\texttt{Retriever-only}.} 
Table~\ref{tab:retireval_results} reports performance across synthetic and converted datasets. Sparse models perform weakest overall, while dense models (\texttt{SFR}, \texttt{Dragon+}, \texttt{BGE-M3}) demonstrate stronger F1 scores across all datasets. Overall, \texttt{Dragon+} achieves the highest average F1 across datasets and cutoffs. This ranking is calculated by averaging the F1 scores computed at each cutoff ($k \in \{1, 3, 5, 7, 10, 20\}$) for each dataset, followed by averaging across all datasets. While the performance gap to the cross-encoders model is not large, due to their higher inference costs, we select \texttt{Dragon+} as the base retriever for subsequent experiments.

\paragraph{\texttt{R+LLM}.}
Building on \texttt{Dragon+}, we next experiment with using LLMs as classifiers to refine the results. Figure~\ref{fig:llm_results} shows the F1 scores of the different \texttt{R+LLM} setups. On the synthetic datasets (\texttt{NEWS-SYNTH} and \texttt{REVIEWS-SYNTH}), applying the LLM as a filter leads to consistent improvements, particularly for \texttt{REVIEWS-SYNTH}, which benefits from the LLM's ability to capture nuanced feedback and critique relations. On the converted datasets (\texttt{NEWS-ECB+} and \texttt{REVIEWS-F1000}), however, the gains are marginal. We attribute this to the mismatch between these datasets and our linking task: \texttt{NEWS-ECB+} focuses on coreference links, while \texttt{REVIEWS-F1000} is derived from \texttt{F1000RD} whose papers are on average longer than papers in \texttt{REVIEWS-SYNTH}, and include domains (biology, medicine) where key evidence appears in figures, limiting LLM effectiveness in our text-only setup. Significance tests (Appendix~\ref{app:sig_tests}) support these observations.
Overall, LLMs add value across domains by capturing relations that are implicit or span multiple lines of reasoning. Thus, while both domains benefit from \texttt{R+LLM}, the nature of their linking tasks determines how large the performance gap over the baseline is. 
Manual error analysis reveals that most failures stem from scattered or ambiguous source sentences, which make fine-grained linking difficult (more details in Appendix~\ref{app:error_analysis}).
Finally, to verify that the improvements of \texttt{R+LLM} are not specific to \texttt{Dragon+}, we also tested the pipeline with other retrievers and observed similar performance gains (see Appendix~\ref{app:retriever_llm} for details).

\paragraph{Choice of LLM}
Across the three LLMs tested (\texttt{Phi-4}, \texttt{Qwen2.5}, and \texttt{GPT-4o}), we find that \texttt{Phi-4} performs on average worse than the others. \texttt{Qwen2.5} is the strongest overall, especially on the synthetic datasets, while \texttt{GPT-4o} slightly outperforms the others on \texttt{NEWS-ECB+}. We attribute this to properties of the dataset itself: \texttt{NEWS-ECB+} links are based on coreferent event mentions, which could be easier for stronger models like \texttt{GPT-4o} to detect. Despite these relative differences, all three models achieve only modest absolute F1 scores on the peer review datasets, highlighting the intrinsic difficulty of sentence-level linking in long, technical documents.
Nevertheless, across all datasets, the best prompting configuration is clear: combining a link description (i.e., a description of what it means for two sentences to be related) with in-context examples leads to the strongest performance. In addition to the prompt content, we find that the way candidate sentences are presented to the LLM matters. In particular, \emph{listwise prompting}, where all top-$k$ candidate sentences are shown together, outperforms \emph{pairwise prompting}, where the LLM considers each candidate in isolation. We hypothesize that listwise prompting allows the LLM to condition its decisions on the global context rather than making them in isolation.

\begin{figure}[!t]
  \includegraphics[width=\linewidth]{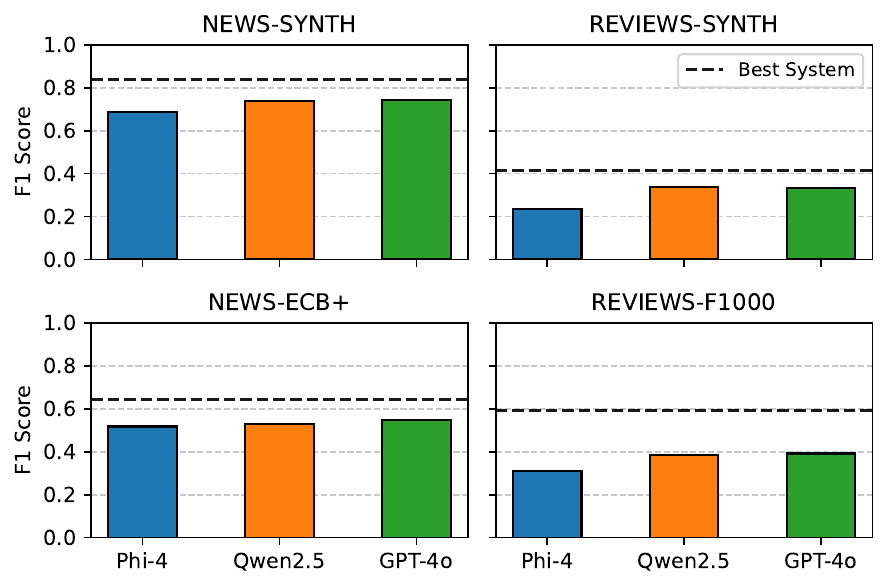}
  \caption{LLM-only ablation. Models were prompted with both full documents (source and target), the specific source sentence, a link description, and in-context examples. While this setup captures more context and task-specific information, it still underperforms compared to the combination of Retriever and LLM. This highlights the importance of retrieval for narrowing the candidate space and reducing distractors, especially in long documents.}
  \label{fig:classification_combination}
\end{figure}

\paragraph{LLM-only ablation.}
To evaluate the necessity of retrieval, we ablate the retriever and use LLMs alone for classification. We use the aforementioned best prompt setup and instruct the models to classify every sentence in the target document as linked or not, (more details in Appendix~\ref{app:llm_only}). As shown in Figure~\ref{fig:classification_combination}, across all datasets, the LLM-only approach underperforms the \texttt{R+LLM} approach, confirming that retrieval is crucial for narrowing the search space and filtering distractors. The performance gap is especially large on the longer more complex peer review documents. For example, \texttt{GPT-4o} reaches only 33.42 and 39.22 F1 on \texttt{REVIEWS-SYNTH} and \texttt{REVIEWS-F1000}, compared to 41.42 and 59.18 with \texttt{R+LLM}. Even on the shorter news articles, the LLM-only setup falls short (e.g., 74.27 vs. 84.02 on \texttt{NEWS-SYNTH}). While LLMs can capture some cross-document signals without retrieval, their high inference cost makes them inefficient as standalone solutions compared to their performance, making retrieval as a first step essential for both efficiency and accuracy.

In summary, our results suggest that the strongest approach for both domains is \texttt{R+LLM} using \texttt{Dragon+} and \texttt{Qwen2.5}, prompting listwise with both a link description and in-context examples.

\section{Assisted labeling}
\label{sec:human_eval}
The final step of our framework applies the best-performing \texttt{R+LLM} approach established on synthetic data to help humans find links in naturally occurring text pairs, and compares it to \texttt{Dragon+} as a baseline in a human evaluation experiment.

\subsection{Setup}
We randomly sample 20 review-paper pairs and news article pairs. \texttt{\textbf{REVIEWS-HE}} consists of short conference papers and their corresponding peer reviews, drawn from the ARR22 subset of the NLPeer dataset. We focus on short papers to avoid long-context limitations during model inference, and select the shortest available review when multiple are present. \texttt{\textbf{NEWS-HE}} comprises article pairs from the SPICED dataset \citep{shushkevich-etal-2024-spiced}, sampled from the \textit{Politics}, \textit{Sports}, and \textit{Culture} categories to ensure topical diversity. In both domains, we segment documents into sentences, followed by manual corrections to improve segmentation accuracy and remove non-content elements such as social media links or boilerplate text. See Appendix~\ref{app:data_prep} for further details on data preparation.
To support the annotation process, we used the INCEpTION platform \citep{klie-etal-2018-inception}, which provides a side-by-side document view and allows annotators to accept or reject pre-highlighted candidate links with access to full sentence and document context; see Appendix~\ref{app:anno_interface} for further details.

\subsection{Protocol}
Annotators were presented with the source and target documents side-by-side, and asked to evaluate candidate links from the source document to the target using domain-specific guidelines (Appendix~\ref{app:annotation_guidelines}). In \texttt{REVIEWS-HE}, up to 8 candidate targets were shown for each source sentence: top-3 from \texttt{R+LLM}, top-3 from \texttt{Dragon+}, and 2 random distractors. In \texttt{NEWS-HE}, which contains shorter documents, the pool was limited to 5 candidates per source sentence: top-2 from \texttt{R+LLM}, top-2 from \texttt{Dragon+}, and one random distractor. To reduce workload, we excluded source sentences that were too short (three words or fewer), as well as explicit links (e.g. Line.X, Figure.Y) which can be resolved trivially. If a candidate target is suggested by both \texttt{R+LLM} and \texttt{Dragon+}, it was only displayed once but is counted toward both during the analysis. Annotators could accept any number of target sentences as linked to a given source, incl. no matches.

\begin{table}[!t]
\centering
\small
\begin{tabular}{lccc}
\toprule
 & \texttt{\textbf{REVIEWS-HE}} & \texttt{\textbf{NEWS-HE}} & \texttt{\textbf{Avg.}} \\
\midrule
\multicolumn{4}{l}{\texttt{Annotation Results}} \\
\midrule
\texttt{R+LLM only} & \texttt{\underline{56.4\%}} & \texttt{\underline{57.0\%}} & \texttt{\underline{56.7\%}} \\
\texttt{Retriever only} & \texttt{42.7\%} & \texttt{17.8\%} & \texttt{30.3\%} \\
\texttt{Both} & \texttt{\textbf{77.7\%}} & \texttt{\textbf{68.6\%}} & \texttt{\textbf{73.1\%}} \\
\texttt{Random} & \texttt{4.3\%} & \texttt{7.7\%} & \texttt{6.0\%} \\
\midrule
\multicolumn{4}{l}{\texttt{Statistics}} \\
\midrule
\texttt{\# Doc. Pairs} & \texttt{20} & \texttt{20} & \\
\texttt{\# Labeled Links} & \texttt{1022} & \texttt{804} & \\
\bottomrule
\end{tabular}
\caption{Acceptance rate between the two annotators on candidate links, broken down by link suggestion method. Also statistics on the number of document pairs and total labeled links for each domain.}
\label{tab:human_eval_link_agreement}
\end{table}

\subsection{Annotation study}
We recruited 15 annotators via Prolific\footnote{\url{https://www.prolific.com/}} and conducted a qualification task using one "gold" document pair with 30 cross-document links labeled by the study authors for each domain (Selection criteria detailed in Appendix~\ref{app:anno_selection}). Annotators were provided with annotation guidelines for each domain. We measured agreement with the gold labels using Cohen’s $\kappa$ \cite{cohen1960coefficient} and selected the two annotators with the highest scores: $\kappa = 0.68$ for \texttt{REVIEWS-HE} and $\kappa = 0.72$ for \texttt{NEWS-HE}, both indicating substantial agreement. The selected annotators then labeled the full evaluation set in two batches of 10 document pairs per domain. After the first batch, we conducted a feedback round to clarify guideline interpretations and highlight common issues, after which the annotation resumed independently. Final inter-annotator agreement across both batches was $\kappa = 0.59$ for \texttt{REVIEWS-HE} and $\kappa = 0.60$ for \texttt{NEWS-HE}, reflecting substantial agreement given the subjective and open-ended nature of the task. On reviews, this exceeds the reported agreement in machine-assisted annotation by \citet{kuznetsov-etal-2022-revise}, despite our use of crowd annotators with minimal training.

\subsection{Results and Analysis}
Table~\ref{tab:human_eval_link_agreement} summarizes the acceptance rates across link suggestion methods. Links suggested by both \texttt{Dragon+} and the \texttt{R+LLM} achieve the highest acceptance rates: \texttt{77.7\%} in \texttt{REVIEWS-HE} and \texttt{68.6\%} in \texttt{NEWS-HE}. This is expected, as mutual agreement between both indicates higher confidence.

More importantly, we observe a consistent pattern across domains: links selected only by the \texttt{R+LLM} approach are accepted at substantially higher rates than those selected only by \texttt{Dragon+}, namely \texttt{56.7\%} vs. \texttt{30.3\%} on average. The contrast is especially pronounced in the news domain (\texttt{57.0\%} vs. \texttt{17.8\%}), where retrieval alone appears less effective. We attribute this to the retriever’s reliance on semantic similarity which fails to capture the nuanced, context-sensitive understanding required for accurate linking. In contrast, LLMs can reason about discourse and topic structure, drawing on prior knowledge to identify implicit connections. This highlights the LLM’s strength as a task-aware filter to refine surface-level retrieval results.

Crucially, this finding echoes our results on synthetic data, where adding LLM classification consistently improved performance. The human evaluation confirms that these gains hold on natural text, demonstrating that the LLM contributes meaningful additional value beyond the retriever’s capabilities. The low acceptance rate for random distractor candidates (\texttt{6.0\%} on average) confirms that the task is non-trivial and that valid cross-document links are unlikely to arise by chance.

Our study results in substantially-sized human-labeled annotated datasets in peer reviews, and news articles, comparable in size to datasets from the literature \citep{kuznetsov-etal-2022-revise, darcy-etal-2024-aries}. While our framework was not designed for precise cost tracking, \citet{darcy-etal-2024-aries} report approx. 30 minutes per pair using fully manual linking on paper edits, in contrast, our annotators completed each full document pair in roughly 15 minutes on average, while annotating 10 times more links. This demonstrates the effectiveness of our candidate filtering in reducing annotation effort. Although the precise savings depend on annotation tools, annotator training, and task familiarity, our results suggest substantial gains in annotation throughput without sacrificing quality.

\subsection{Estimating True Recall}
\label{est_true_recall}
Similar to other assisted labeling settings, our setup does not provide an estimate of true recall, since only candidate links were annotated. To measure true recall, we conducted a supplementary experiment where we exhaustively labeled \textit{all} links on a small subset of approx.\ 10\% of \texttt{REVIEWS-HE} and \texttt{NEWS-HE}, resulting in 102 links for reviews and 80 for news. We use this subset to estimate the true performance of the link suggestion approaches. As Table~\ref{tab:true_recall_results} demonstrates, across both domains, \texttt{R+LLM} substantially improves recall and precision over retrieval-only baselines. As expected for an intersection strategy, the \texttt{Both} method demonstrates a drop in recall, and consequently does not outperform \texttt{R+LLM} on overall F1. This demonstrates that while \texttt{Both} can serve as a conservative high-confidence filter in annotation settings, it is not a balanced solution when considering recall. Note that in the main human study (Table \ref{tab:human_eval_link_agreement}), higher acceptance rates are due to annotating top-$k$ candidates only, whereas Table \ref{tab:true_recall_results} measures full-corpus metrics. We provide details and additional analysis in Appendix~\ref{app:manual_recall}. In sum, LLMs capture links that retrieval alone misses, and validates our framework’s effectiveness in high-recall scenarios.

\begin{table}[!t]
\centering
\small
\begin{tabular}{lccc}
\toprule
& \texttt{Recall} & \texttt{Precision} & \texttt{F1} \\
\midrule
\multicolumn{4}{l}{\textbf{\texttt{NEWS}}} \\
\midrule
\texttt{R+LLM}     & \texttt{\textbf{0.77}} & \texttt{\textbf{0.93}} & \texttt{\textbf{0.82}} \\
\texttt{Retriever} & \texttt{0.57} & \texttt{0.70} & \texttt{0.61} \\
\texttt{Both}      & \texttt{0.54} & \texttt{0.76} & \texttt{0.61} \\
\texttt{Random}    & \texttt{0.02} & \texttt{0.05} & \texttt{0.02} \\
\midrule
\multicolumn{4}{l}{\textbf{\texttt{REVIEWS}}} \\
\midrule
\texttt{R+LLM}     & \texttt{\textbf{0.59}} & \texttt{\textbf{0.62}} & \texttt{\textbf{0.55}} \\
\texttt{Retriever} & \texttt{0.28} & \texttt{0.26} & \texttt{0.25} \\
\texttt{Both}      & \texttt{0.26} & \texttt{0.37} & \texttt{0.27} \\
\texttt{Random}    & \texttt{0.00} & \texttt{0.00} & \texttt{0.00} \\
\bottomrule
\end{tabular}
\caption{True recall estimation results under exhaustive evaluation. While ``Both'' achieves higher precision, its recall is substantially reduced, limiting its overall F1. Metrics are macro-averaged across source sentences.}
\label{tab:true_recall_results}
\end{table}

\section{Conclusion}

High annotation cost and diversity of link types hinder progress in cross-document linking. We present a domain-agnostic framework for sentence-level linking that bootstraps annotation with no labeled data upfront. By generating synthetic linked documents, we enable automatic evaluation of linking approaches and identify the best-performing ones for assisted human annotation.

Applied in peer review and news, our framework shows that combining retrieval with LLM-based classification yields high-quality links, doubling the recall and precision of retrieval alone. It generalizes across domains, lowers annotation effort, and produces reusable datasets. Our results highlight the framework’s potential for scalable, real-world cross-document analysis.

\section*{Ethical considerations}

Our work does not carry substantial additional risks and contributes to better technological support for many socially relevant application areas such as academic quality control, journalism, fake news detection and propaganda analysis. 
As with any AI technology, we call for additional testing and oversight if the proposed method is deployed in a sensitive domain such as law, medicine or critical infrastructure. Some general ethical risks include bias due to the use of synthetic data, lower performance in under-represented languages and domains, and dual use, for example using links between news articles and social media commentary to spot dissent. Moreover, if the generated links are used without evaluation, then incorrect links could lead to potential harms like misinformation (news domain) and wrong assumptions (peer-review domain). The participants of the synthetic data validation study in the news domain contributed on a voluntary basis; the participants of the synthetic data validation study on peer reviews conducted it as a part of their employment at the authors' institution. The crowdworkers on Prolific have been fairly compensated with 13 Euro per hour. The NLPeer dataset has a \texttt{CC BY 4.0} license, and we release our corresponding derivative dataset under the same license. The ECB+ corpus has a \texttt{CC BY 3.0} license which allows us to release the corresponding derivative dataset under \texttt{CC BY 4.0} license. The SPICED dataset has a \texttt{CC BY 4.0} license, but the licensing for its source material (news) is unclear. To work around this, we release a script for reconstructing the dataset along with our added data. All datasets were used according to their intended research purpose. 

\section*{Limitations}
\label{sec:limitations}

\paragraph{Generality vs.\ domain specificity.}
We propose a general framework for linking annotation and demonstrate its effectiveness across two distinct domains. We note that while the overall framework is agnostic to domain and link type, some steps (data generation, prompting, evaluation) need to be instantiated with domain-specific configurations (e.g., link definitions, sampling strategies, prompt templates). We view this as a modest cost compared to the human effort required for fully manual annotation, as it balances general applicability with minimal task-specific tuning. Domain-specific characteristics (e.g., subjective critique in reviews vs.\ factual alignment in news) can affect how well a general approach performs out-of-the-box. Thus, more specialized models or task formulations might yield stronger results in single-domain settings. Future work could explore adaptive modules or domain-specialized tuning within our general framework, as well as evaluate the applicability of our framework in further domains.

\paragraph{Synthetic data.}
Our approach to generating synthetic data relies on prompting an LLM to produce source documents conditioned on a target document. While results show this is effective, coverage could be improved through more targeted generation strategies—such as prompting for individual sentence-level links or applying multi-pass generation. That could also improve the realism of the links as discussed in Section~\ref{sec:validation}. We also note that small human-annotated gold sets could in principle support retrieval model selection, but as shown in Section~\ref{est_true_recall} creating them is costly and often impractical at scale. Our synthetic approach is designed for the common case where no labeled data exists upfront; when domain-specific gold data is available, it can be incorporated as a complementary signal in the evaluation stage. Finally, LLM hallucinations can lead to synthetic data that does not follow the intended goal. While we ensure data quality through human validation, targeted evaluation of LLM hallucinations in synthetic linking data generation lies beyond our scope and is left for future work.

\paragraph{Model coverage and training.}
Due to cost constraints, we trade off the number of LLMs against the depth of evaluation for each. We prioritize coverage of representative retrievers and LLMs, rather than exhaustive prompt variants or multiple runs. Expanding these dimensions would offer deeper insight into robustness and prompt sensitivity.
We do not fine-tune smaller models like BERT, as our goal is fast iteration without task-specific supervision. Although we could fine-tune on our generated synthetic data, this type of training on synthetic data is known to limit generalization in subjective tasks \citep{li-etal-2023-synthetic}.

\paragraph{Document modalities.}
Our framework currently operates on text only. However, in domains like scientific writing, non-textual elements (e.g., tables, figures) often contain essential linkable information. Supporting multi-modal linking is an important direction for future extensions of our method.

\paragraph{Link coverage.}
A core limitation of assisted annotation is that recall cannot be fully measured without exhaustive manual labeling. This is a known challenge in linking and retrieval tasks more broadly: any system that filters or ranks candidate links will inevitably leave some valid links unexamined. To mitigate this, we first tune the retrieval cutoff on synthetic data to maximize recall, ensuring that high-coverage candidates are passed to the LLM. Second, we conduct a targeted manual annotation study with full link coverage to estimate true recall on a subset of document pairs.

\section*{Acknowledgments}
This work has been co-funded by the German Federal Ministry of Research, Technology and Space (BMFTR) under the promotional reference 01ZZ2314H (GeMTeX), and by the European Union (ERC, InterText, 101054961). Views and opinions expressed are however those of the author(s) only and do not necessarily reflect those of the European Union or the European Research Council. Neither the European Union nor the granting authority can be held responsible for them. We gratefully acknowledge the support of Microsoft with a grant for access to OpenAI GPT models via the Azure cloud (Accelerate Foundation Model Academic Research). We thank Simone Balloccu for insightful feedback, and Thy Thy Tran, Hassan Soliman and Shivam Sharma for helpful comments on an earlier draft of this paper.

\bibliography{anthology,custom}

\appendix
\section{Group-level link analysis}
\label{app:group_links}

To assess the extent of group-level relations, we analyzed clusters of adjacent links, defined as sets of two or more links where source and/or target sentences are consecutive. Such clusters approximate larger argumentative or narrative units.

\begin{table*}[ht]
\centering
\small
\begin{tabular}{lccccc}
\toprule
\textbf{Domain} & \textbf{Dataset} & \textbf{\#Pairs} & \textbf{\#Links} & \textbf{\%Links in Clusters} & \textbf{Mean / Max Size} \\
\midrule
News & Synthetic & 346 & 2323 & 70\% & 3.1 / 7 \\
Reviews & Synthetic & 200 & 2181 & 44\% & 2.4 / 7 \\
News & Human & 20 & 296 & 54\% & 3.1 / 12 \\
Reviews & Human & 20 & 334 & 21\% & 2.2 / 4 \\
\bottomrule
\end{tabular}
\caption{Analysis of clusters of consecutive links. Group-level structures are frequent, especially in news.}
\label{tab:group_link_stats}
\end{table*}
The results in Table \ref{tab:group_link_stats} confirm that group-level structures are common-particularly in news, where shorter documents promote multi-sentence alignments. Importantly, the analysis also shows that sentence-level links compose naturally into such clusters, providing atomic building blocks for broader structures. Future extensions of our framework may incorporate explicit span-level or hierarchical modeling.

\section{Datasets Conversion}
\label{app:data_conversion}
In \textbf{\texttt{NEWS-ECB+}}, for each document pair, we label a sentence pair as linked if they contain event mentions in the same cluster. While this conversion provides a useful signal, it introduces two key limitations: (1) the resulting labels are incomplete, as only coreferent events are annotated, and (2) some event mentions occur in headlines or titles, which leads to spurious links from titles to many other sentences. In
\textbf{\texttt{REVIEWS-F1000}}, the authors didn't adjudicate the annotations, so we only select examples where both expert annotators agreed on the label. This reduces the number of links we can use, but ensures we have a comparable dataset to \textbf{\texttt{REVIEWS-SYNTH}}.

\section{Quantitative Criteria}
\label{app:quan_criteria}
\begin{figure*}[ht]
  \centering
  \begin{subfigure}[t]{0.95\textwidth}
    \centering
    \includegraphics[width=\linewidth]{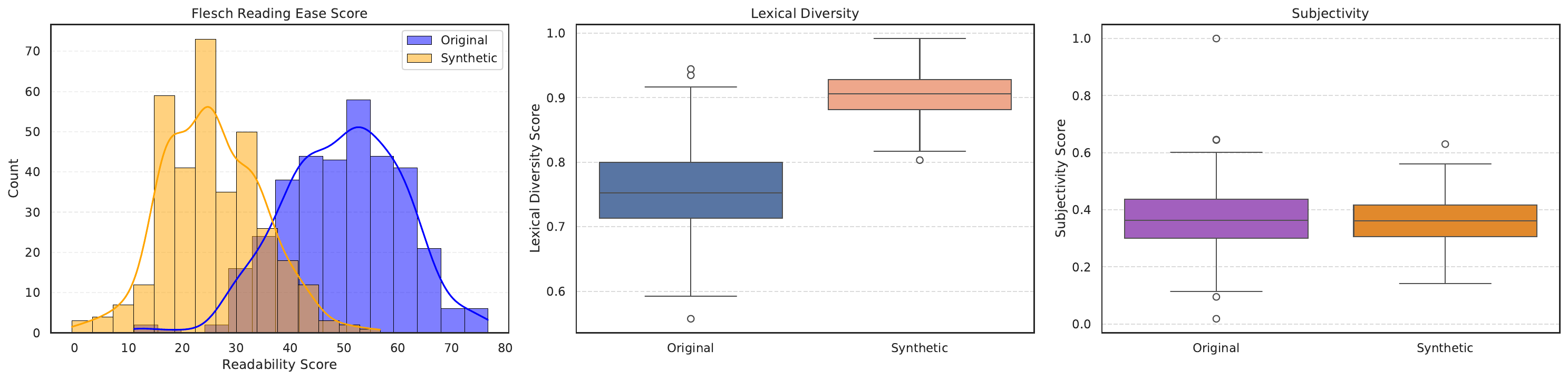}
    \caption{Quantitative comparison of synthetic news to natural news data.}
    \label{fig:synth_news_comparison}
  \end{subfigure}
  
  \vspace{1em}

  \begin{subfigure}[t]{0.95\textwidth}
    \centering
    \includegraphics[width=\linewidth]{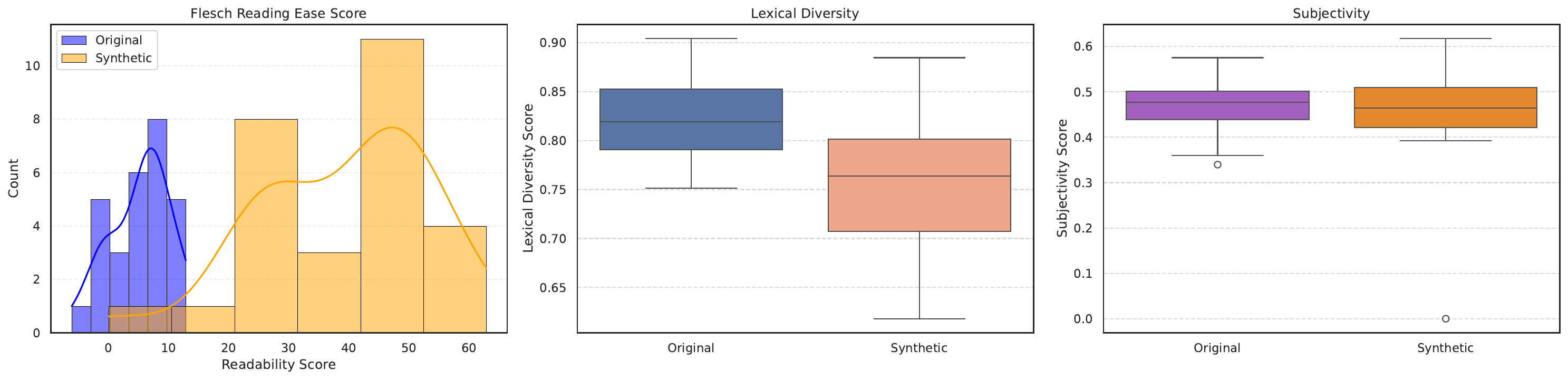}
    \caption{Quantitative comparison of synthetic reviews to natural reviews.}
    \label{fig:synth_reviews_comparison}
  \end{subfigure}
  
  \caption{Comparison of synthetic and natural documents in terms of stylistic and structural metrics, across two domains.}
  \label{fig:synthetic_vs_real}
\end{figure*}

To assess how closely our synthetic documents resemble natural ones, we selected three document-level metrics that capture distinct stylistic dimensions: (1) lexical diversity, as a proxy for richness of vocabulary; (2) subjectivity\footnote{\href{https://huggingface.co/cffl/bert-base-styleclassification-subjective-neutral}{https://huggingface.co/cffl/bert-base-styleclassification-subjective-neutral}}, to assess differences in tone or stance; and (3) lexical complexity, using the Flesch-Kincaid Reading Ease score, to estimate linguistic difficulty. These criteria were chosen to reflect qualities that may affect linkability and document coherence, while remaining agnostic to content. Figure~\ref{fig:synthetic_vs_real} plots the results for synthetic and natural documents in both domains.

\section{Rationale for Evaluation Criteria}
\label{app:human_val}
We selected domain-specific evaluation criteria to reflect both general text quality and the particular requirements of our linking task.
For \texttt{NEWS-SYNTH}, we assessed:
\begin{itemize}
    \item \textbf{Fluency} assesses grammatical correctness and naturalness of the language, ensuring the text is readable.
    \item \textbf{Coherence} measures logical flow and topic consistency, which is critical for linking tasks that depend on understanding article structure and progression.
    \item \textbf{Realism} evaluates whether the article could plausibly have been written by a human journalist, and ensure the generated article is not clearly AI-generated.
    \item \textbf{Specificity} checks for the inclusion of concrete facts (e.g., names, dates, events, etc.) to ensure the AI-generated review is not general and vague.
\end{itemize}

\begin{figure*}[ht]
  \centering
  \begin{subfigure}[b]{0.48\linewidth}
    \includegraphics[width=\linewidth]{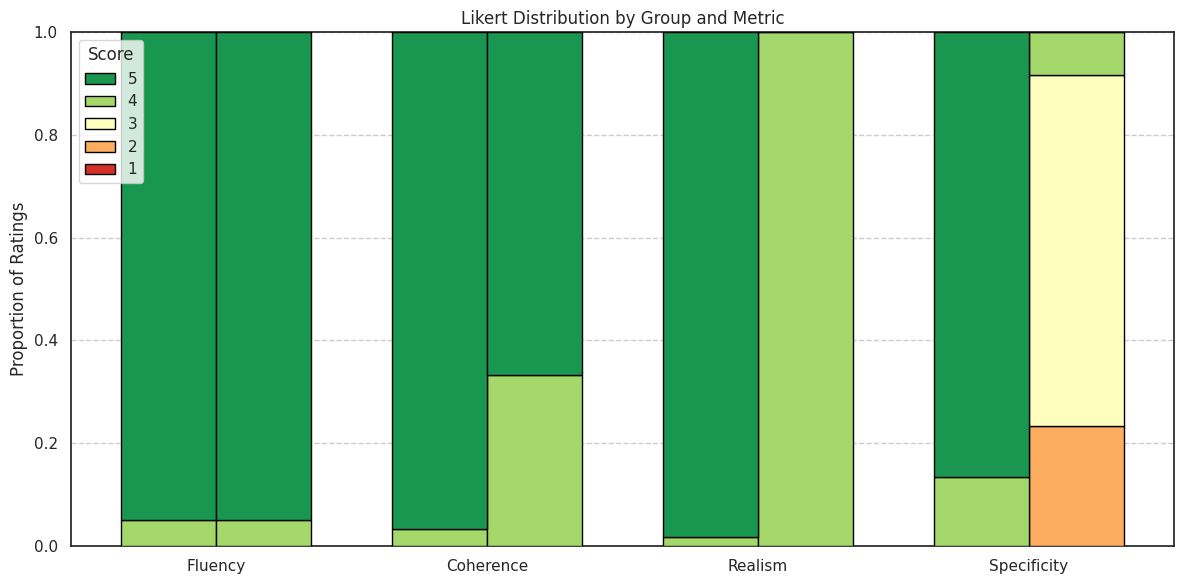}
    \caption{\texttt{NEWS-SYNTH} Human Validation.}
    \label{fig:synth_news_annos_discrete}
  \end{subfigure}
  \hfill
  \begin{subfigure}[b]{0.48\linewidth}
    \includegraphics[width=\linewidth]{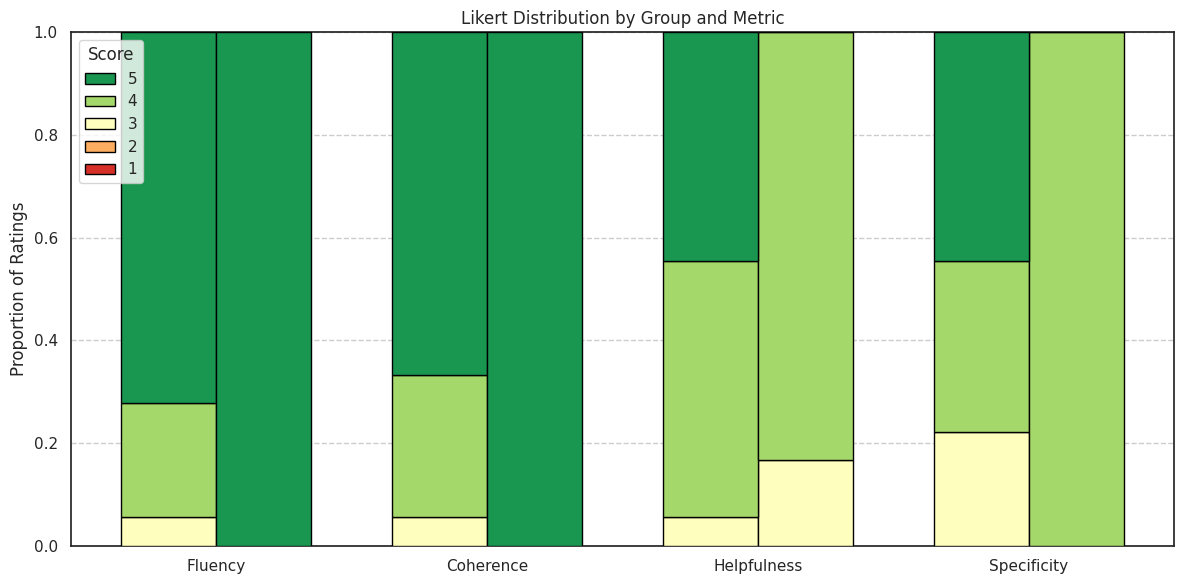}
    \caption{\texttt{REVIEWS-SYNTH} Human Validation.}
    \label{fig:synth_reviews_annos_discrete}
  \end{subfigure}
  \caption{Human evaluation results for \texttt{NEWS-SYNTH} (left) and \texttt{REVIEWS-SYNTH} (right).}
  \label{fig:synth_combined_eval}
\end{figure*}

For \texttt{REVIEWS-SYNTH}, we assessed:
\begin{itemize}
    \item \textbf{Fluency} and \textbf{coherence}, as above, ensure the review reads naturally and follows a logical structure.
    \item \textbf{Helpfulness} replaces realism in this context, measuring whether the review provides meaningful feedback that could assist authors, which is central to the function of peer reviews.
    \item \textbf{Specificity} captures whether the review addresses concrete elements of the paper, such as specific sections, claims, or experiments, helping differentiate generic summaries from insightful critique.
\end{itemize}
These criteria were chosen to ensure that our synthetic datasets maintain linguistic quality and task relevance, even if they do not replicate all the nuanced characteristics of human writing. Figure~\ref{fig:synth_combined_eval} illustrates detailed validation results.

\section{Retrievers' Results}
\label{app:detailed_results}
Tables \ref{tab:news_ecb}, \ref{tab:news_synth}, \ref{tab:reviews_synth} and \ref{tab:reviews_F1000} provide the full retrieval results for all models for all datasets individually, broken down by cutoff values ($k \in \{1, 3, 5, 7, 10, 20\}$).

\section{Statistical significance testing}
\label{app:sig_tests}

We conducted bootstrap resampling (10k iterations) to test whether improvements from adding LLM classification (\texttt{R+LLM}) over the best retriever baseline (\texttt{Dragon+}) are statistically significant. Results are shown in Table~\ref{tab:significance_results}.

\begin{table*}[ht]
\centering
\small
\begin{tabular}{lccccc}
\toprule
\textbf{Dataset} & \textbf{Retriever F1} & \textbf{R+LLM F1} & \textbf{$\Delta$F1} & \textbf{95\% CI} & \textbf{p-value} \\
\midrule
NEWS-SYNTH    & 0.74 & 0.84 & +0.096 & [0.045, 0.140] & <0.001 \\
REVIEWS-SYNTH & 0.25 & 0.41 & +0.159 & [0.127, 0.188] & <0.001 \\
NEWS-ECB+     & 0.55 & 0.57 & +0.015 & [-0.011, 0.042] & 0.72 \\
REVIEWS-F1000 & 0.59 & 0.59 & -0.002 & [-0.020, 0.032] & 0.22 \\
\bottomrule
\end{tabular}
\caption{Bootstrap significance test of F1 improvements from R+LLM over Dragon+.}
\label{tab:significance_results}
\end{table*}

The results confirm that improvements on the synthetic datasets are statistically significant, whereas gains on the converted datasets are not, which aligns with our earlier discussion that these datasets are less aligned with our task definition.

\section{Error analysis}
\label{app:error_analysis}
To better understand the limitations of our approach, we manually analyzed 100 failure cases across datasets. We identified three frequent patterns:
\begin{enumerate}
    \item Scattered target sentences: this happens when the target sentences are not concentrated in a specific location in the target document, but are spread out. This is especially a problem in the reviews domain where some review sentences address something in the paper that’s presented listwise over multiple pages (e.g. steps of a methodology). This problem is present in the news domain too, but due to the shorter document sizes, it’s less prominent.
    \item Ambiguous source sentences: Some source sentences lack sufficient signals or specificity for reliable linking (e.g., short, vague.) Similar to the point above, this is a problem in the reviews domain where reviews are shorter documents with more condensed sentences. However, even in the news domain, depending on the article’s source and its editing style, some sentences can lack sufficient signals necessary for linking.
    \item Too many distractors: Because we use the retriever to reduce the search space, we end up passing a list of very semantically similar target sentences to the LLM. In the reviews domain, this is less of a problem, because the more semantically similar sentences are more concentrated in the same sections on the paper, but this in itself leads to problem 1 mentioned above. In the news domain, due to the shorter documents, there are similar sentences that get passed to the LLM requiring more reasoning to filter them out.
\end{enumerate}

\section{R+LLM across different retrievers}
\label{app:retriever_llm}

To test whether the benefits of LLM-classification depend on the choice of base retriever, we applied the \texttt{R+LLM} step (Qwen2.5, listwise prompting with description + examples) to four additional retrieval models: \texttt{SFR}, \texttt{ms\_marco\_MiniLM}, \texttt{BM25}, and \texttt{BGE-M3-sparse}. Results are shown in Table~\ref{tab:retriever_llm_results}.

\begin{table*}[ht]
\centering
\small
\begin{tabular}{lcccc}
\toprule
\textbf{Dataset} & \textbf{Retriever} & \textbf{Retriever Only} & \textbf{R+LLM} & \textbf{Diff} \\
\midrule
NEWS-ECB+ & SFR & 58.19 & 53.88 & -4.31 \\
          & ms\_marco\_MiniLM & 54.63 & 54.40 & -0.23 \\
          & BM25 & 45.03 & 53.44 & +8.41 \\
          & BGE-M3-sparse & 48.30 & 53.65 & +5.35 \\
          & \textit{Dragon+} & \textit{55.36} & \textit{56.84} & +1.48 \\
\midrule
NEWS-SYNTH & SFR & 70.04 & 78.18 & +8.14 \\
           & ms\_marco\_MiniLM & 77.69 & 80.56 & +2.87 \\
           & BM25 & 61.78 & 76.63 & +14.85 \\
           & BGE-M3-sparse & 67.56 & 77.61 & +10.05 \\
           & \textit{Dragon+} & \textit{74.38} & \textit{84.02} & +9.64 \\
\midrule
REVIEWS-SYNTH & SFR & 23.09 & 36.16 & +13.07 \\
              & ms\_marco\_MiniLM & 27.60 & 37.12 & +9.52 \\
              & BM25 & 22.21 & 35.53 & +13.32 \\
              & BGE-M3-sparse & 25.79 & 38.12 & +12.33 \\
              & \textit{Dragon+} & \textit{25.48} & \textit{41.42} & +15.94 \\
\midrule
REVIEWS-F1000 & SFR & 42.44 & 42.07 & -0.37 \\
              & ms\_marco\_MiniLM & 61.03 & 57.38 & -3.65 \\
              & BM25 & 54.75 & 59.14 & +4.39 \\
              & BGE-M3-sparse & 54.11 & 57.97 & +3.86 \\
              & \textit{Dragon+} & \textit{58.73} & \textit{58.56} & -0.17 \\
\bottomrule
\end{tabular}
\caption{Performance of R+LLM using Qwen2.5 across different retrievers. Scores are F1 on synthetic and converted datasets.}
\label{tab:retriever_llm_results}
\end{table*}

On the synthetic datasets, the R+LLM setup yields consistently positive and similarly sized improvements on REVIEWS-SYNTH ($\approx$+9–16 F1) compared to the other synthetic dataset NEWS-SYNTH ($\approx$+8–14 F1). On converted datasets (NEWS-ECB+ and REVIEWS-F1000), improvements are marginal, which we attribute to task mismatch (Section~\ref{sec:data}). Interestingly, on REVIEWS-F1000, BM25 shows the largest positive improvement among the retrievers when combined with the LLM, also achieving the highest F1 in the R+LLM setup. This matches our explanation that BM25’s lexical matches are particularly helpful in this biomedical/technical domain due to domain-specific terminology in biology and medicine. On the other datasets BM25 starts from a lower base performance than the dense retrievers, hence the LLM has more room to correct errors, yielding larger relative gains.

\section{LLM Prompting Techniques}
\label{app:llm_prompts}
We created two prompt templates corresponding to the \textit{pairwise} and \textit{listwise} classification setups (Figures~\ref{fig:prompt_llm_pairwise} and~\ref{fig:prompt_llm_listwise}). Each prompt includes a system message that frames the task: determining whether a candidate target sentence from one document is related to a source sentence from another document, using full document context. In the pairwise setup, the model classifies one sentence pair at a time; in the listwise setup, it classifies a ranked list of candidate targets in a single pass. Both formats support four prompting configurations: no guidance, link description only, in-context examples only, and both. The prompts are designed to output binary classifications and return structured JSON outputs, facilitating scalable evaluation across top-$k$ retrieved candidates. For \texttt{Phi-4} and \texttt{Qwen2.5}, we used vLLM \cite{DBLP:conf/sosp/KwonLZ0ZY0ZS23} with the structured outputs function to ensure output consistency. The experiments with the open-source models were run on A100 GPUs. For \texttt{GPT-4o}, we used the OpenAI API with structured outputs too. For all models, we set \texttt{temperature = 0.3} and \texttt{top-p = 0.9}.

\begin{figure}[t]   
\centering
\begin{tcolorbox}[colback=gray!5, colframe=gray!80, fontupper=\ttfamily, title=LLM Classification Prompt (Pairwise), width=\linewidth]
\footnotesize
\textbf{System Message:}\\
You are an AI assistant specialized in evaluating sentence relations.\\
You will get two related documents, along with a sentence from Document 1 (source) and a sentence from Document 2 (target). Your task is to determine if the target sentence is related to the source sentence.\\

Prompt Format:\\
- Full Document 1: [text of Document 1] \\
- Full Document 2: [text of Document 2] \\
- Source Sentence from Document 1: [source sentence] \\
- Target Sentence from Document 2: [target sentence] \\

Prompt Variants:\\
- \textbf{Mode 1 (No Guidance):} No additional information; model is directly instructed to decide if the pair is related.\\
- \textbf{Mode 2 (Examples Only):} A few positive example sentence pairs are provided to guide the model.\\
- \textbf{Mode 3 (Description Only):} A link description is provided to define what counts as a "link."\\
- \textbf{Mode 4 (Description + Examples):} Both the link description and examples are included.\\

Response Format:\\
- A JSON object of the form: \texttt{\{"related": true\}} or \texttt{\{"related": false\}}\\
\end{tcolorbox}
\caption{Prompt template for LLM-based pairwise sentence classification.}
\label{fig:prompt_llm_pairwise}
\end{figure}

\begin{figure}[t]
\centering
\begin{tcolorbox}[colback=gray!5, colframe=gray!80, fontupper=\ttfamily, title=LLM Classification Prompt (Listwise), width=\linewidth]
\footnotesize
\textbf{System Message:}\\
You are an AI assistant specialized in evaluating sentence relations.\\
You will get two related documents, along with a sentence from Document 1 (source) and a list of sentences from Document 2 (targets). The targets are ranked based on their similarity to the source sentence. Your task is to determine for each target sentence if it is related to the source sentence. This will help filter out irrelevant sentences and improve the quality of the ranked sentences.\\

Prompt Format:\\
- Document 1: [text of Document 1]\\
- Document 2: [text of Document 2]\\
- Source Sentence from Document 1: [source sentence]\\
- Ranked Target Sentences from Document 2 (Sentence\_ID: Sentence\_text):\\
\quad \texttt{0: "..."}\\
\quad \texttt{1: "..."}\\
\quad ...\\

Prompt Variants:\\
- \textbf{Mode 1 (No Guidance):} Direct classification with no additional context.\\
- \textbf{Mode 2 (Examples Only):} Positive examples are provided before classification.\\
- \textbf{Mode 3 (Description Only):} A definition of what constitutes a "related" sentence is given.\\
- \textbf{Mode 4 (Description + Examples):} Both the description and examples are included.\\

Response Format:\\
- A JSON object with sentence IDs as keys and \texttt{true} or \texttt{false} as values.\\
\quad e.g., \texttt{\{"0": true, "1": false, "2": true\}}\\
\end{tcolorbox}
\caption{Prompt template for LLM-based listwise sentence classification.}
\label{fig:prompt_llm_listwise}
\end{figure}

\section{LLM-only Ablation}
\label{app:llm_only}
Figure~\ref{fig:prompt_llm_classification} shows the prompt template used in the LLM-only setup, where the model classifies all sentences in the target document without prior retrieval. The prompt includes full document context and a single source sentence, along with a link description, and in-context examples, and instructs the LLM to evaluate each target sentence for its relevance. As in the other setups, we support four prompt configurations: no guidance, with in-context examples, with a link description, and with both. We used structured outputs here too, with \texttt{temperature = 0.3} and \texttt{top-p = 0.9}

\begin{figure}[t]
\centering
\begin{tcolorbox}[colback=gray!5, colframe=gray!80, fontupper=\ttfamily, title=LLM-only Classification Prompt, width=\linewidth]
\footnotesize
\textbf{System Message:}\\
You are an AI assistant specialized in evaluating sentence relations.\\
You will get two related documents, along with a sentence from Document 1 (source). Your task is to determine for each sentence in the target document (Document 2) if it is related to the source sentence. This will help filter out irrelevant sentences.\\

Prompt Format:\\
- Document 1: [text of Document 1]\\
- Document 2: [text of Document 2]\\
- Source Sentence from Document 1: [source sentence]\\

Prompt Variants:\\
- \textbf{Mode 1 (Minimal):} No additional context—just classify based on the input documents and source sentence.\\
- \textbf{Mode 2 (Examples Only):} A few examples of positive sentence pairs are provided.\\
- \textbf{Mode 3 (Description Only):} A link description is provided to define what counts as a link.\\
- \textbf{Mode 4 (Description + Examples):} Both the link description and in-context examples are included.\\

Response Format:\\
- A JSON object where each key is a target sentence ID and each value is either \texttt{true} or \texttt{false}.\\
\quad e.g., \texttt{\{"11": true, "4": false, "9": true\}}\\
- Ensure that every sentence in Document 2 is classified.\\
\end{tcolorbox}
\caption{Prompt template for LLM-only sentence classification across an entire target document.}
\label{fig:prompt_llm_classification}
\end{figure}

\section{Data Preparation for Human Evaluation}
\label{app:data_prep}
For \texttt{REVIEWS-HE}, we randomly sampled 20 paper-review pairs from the ARR22 subset of NLPeer from the shortest 100 ones to minimize long-context issues during model inference. In cases with multiple reviews, we selected the shortest available. Peer reviews were segmented into sentences using spaCy \footnote{model: en\_core\_web\_md} \citep{Honnibal_spaCy_Industrial-strength_Natural_2020} and manually corrected for sentence boundary errors. The paper texts came pre-segmented via NLPeer.
For \texttt{NEWS-HE}, we sampled 20 article pairs from the SPICED dataset across the \textit{Politics}, \textit{Sports}, and \textit{Culture} categories to ensure diversity. The SPICED dataset is a multi-document summarization dataset that pairs news articles of different sources that talk about the same topic or event, so the document pairs are naturally related, and suitable for our task.

\section{Annotation interface}
\label{app:anno_interface}
\begin{figure*}[t]
  \centering
  \fbox{\includegraphics[width=\linewidth]{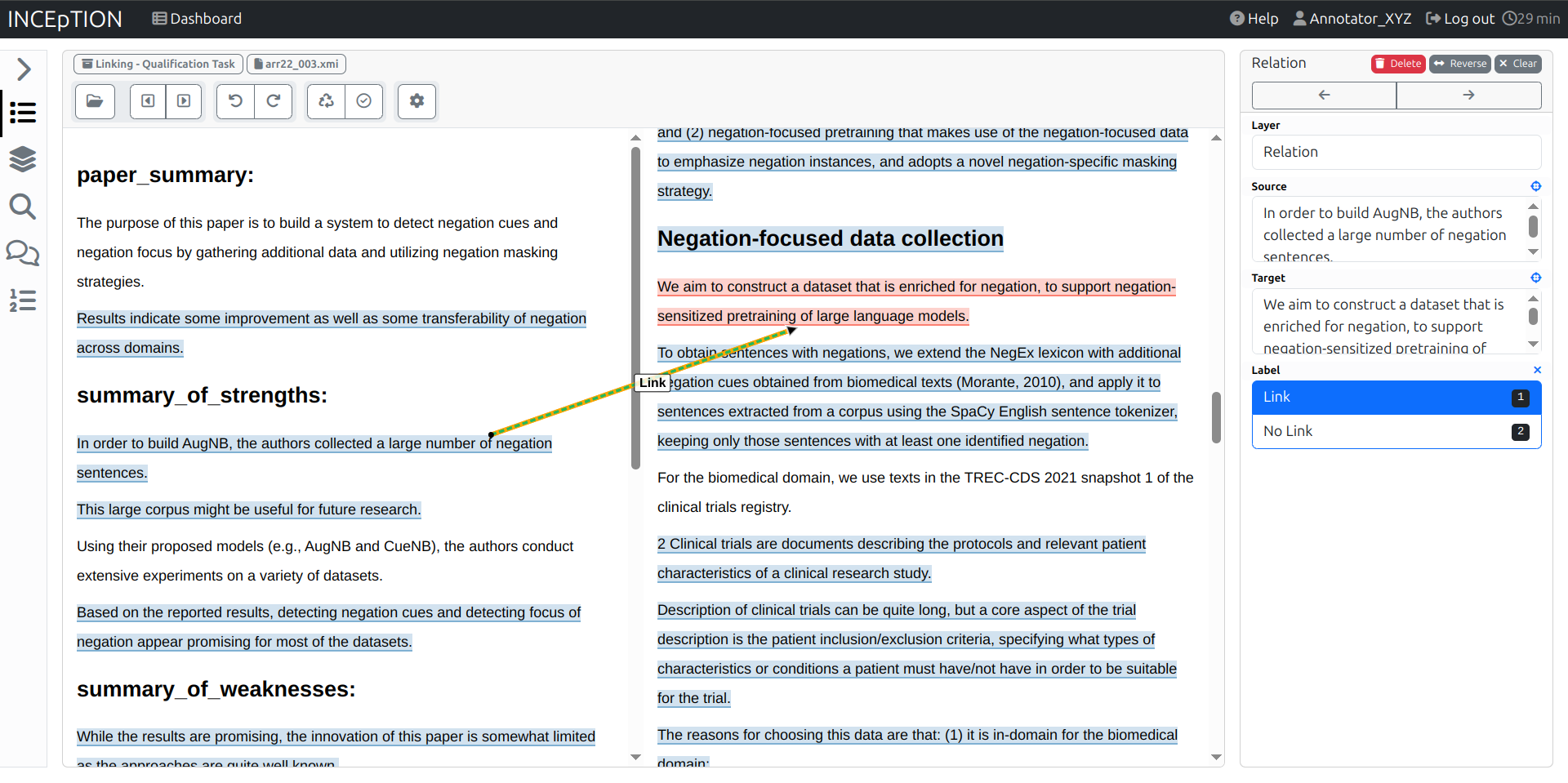}}
  \caption{Annotation Interface.}
  \label{fig:annotation_interface}
\end{figure*}

We used the INCEpTION annotation platform \citep{klie-etal-2018-inception}, which supports side-by-side viewing of two documents and interactive annotation (Figure \ref{fig:annotation_interface}). Source sentences, along with their candidate target sentences were pre-highlighted in the interface, along with visual cues, in the form of arrows pointing from the source sentence to each target sentence, to reinforce the linking aspect of the task. Annotators could view the full sentence and surrounding context and were asked to accept or reject each candidate. Candidates from both the retriever and \texttt{R+LLM} approach were visually indistinguishable and shown in the normal order they show up in the document.

\section{Annotator Selection Criteria}  
\label{app:anno_selection}
Annotators were screened based on the following eligibility criteria: Native English speakers, residing in United States, United Kingdom, Canada, Australia, or Ireland, with at least a Master's or PhD degree in Computer Science, a Prolific approval rate of at least 90\%, and had completed a minimum of 30 prior studies on the platform.

\section{Agreement Trends and Annotation Feedback}
\label{app:agreement_feedback}
We observed a learning curve in annotator agreement. In \texttt{REVIEWS-HE}, agreement started relatively low, improved steadily toward the middle of the batch, and declined slightly toward the end, likely due to annotator fatigue. In \texttt{NEWS-HE}, agreement rose more gradually and plateaued at a slightly lower level overall. First-batch inter-annotator agreement was $\kappa = 0.55$ for \texttt{REVIEWS-HE} and $\kappa = 0.54$ for \texttt{NEWS-HE}, indicating moderate agreement despite the subjective nature of the task.

After the first batch, we reviewed disagreement patterns. One annotator tended to reject links that relied on broader document context, favoring self-contained links. The other applied stricter interpretations of the guidelines, occasionally rejecting links that only partially aligned. Based on these observations, we gave targeted feedback, clarified borderline cases, and encouraged consistent application of the criteria. This led to improved agreement in the second batch ($\kappa = 0.62$ for \texttt{REVIEWS-HE}, $0.64$ for \texttt{NEWS-HE}). However, this could also be attributed to the annotators getting used to the task, and thus understanding the guidelines better.

\section{Annotation Guidelines}
\label{app:annotation_guidelines}
Table \ref{tab:linking_criteria_reviews} and \ref{tab:linking_criteria_news} contain the criteria provided to the annotators to decide whether a sentence pair should be labeled as linked or not. Because the concept of a "link" can be broad, we provided domain-specific examples to clarify what qualifies as a meaningful connection in the context of peer reviews and news.

\begin{table}[t]
\centering
\footnotesize
\setlength{\tabcolsep}{6pt}
\renewcommand{\arraystretch}{1.4}
\begin{tabular}{|>{\raggedright\arraybackslash}p{0.45\linewidth}|>{\raggedright\arraybackslash}p{0.45\linewidth}|}
\hline
\rowcolor{gray!20}
\textbf{Link} & \textbf{No Link} \\
\hline
The review sentence critiques or praises the same claim or result as the paper sentence. & The review mentions grammar, structure, or other writing aspects not discussed in the candidate sentence. \\
\hline
The review analyzes or extends a method/result introduced in the paper sentence. & The review proposes future work or experiments not mentioned in the paper sentence. \\
\hline
The review addresses the same technical method, component, or result. & The review discusses different topics or experiments from the candidate sentence. \\
\hline
\end{tabular}
\caption{Linking Criteria for Peer Reviews}
\label{tab:linking_criteria_reviews}
\end{table}

\begin{table}[t]
\centering
\footnotesize
\setlength{\tabcolsep}{6pt}
\renewcommand{\arraystretch}{1.4}
\begin{tabular}{|>{\raggedright\arraybackslash}p{0.45\linewidth}|>{\raggedright\arraybackslash}p{0.45\linewidth}|}
\hline
\rowcolor{gray!20}
\textbf{Link} & \textbf{No Link} \\
\hline
Both sentences refer to the same event or topic. & Sentences refer to different events, even if they happen at the same place/time. \\
\hline
One sentence is a follow-up or elaboration of the other. & The target sentence provides unrelated opinion, editorial, or background. \\
\hline
One is a paraphrase or restatement of the other (even if phrased differently). & The anchor and target describe distinct stories with no direct overlap. \\
\hline
\end{tabular}
\caption{Linking Criteria for News}
\label{tab:linking_criteria_news}
\end{table}

\section{Synthetic Data Generation Prompts}
\label{app:synth_data_gen}
To generate synthetic peer reviews and news articles, we designed detailed prompting templates that reflect realistic domain practices while introducing controlled variation. The peer review prompt (Figure~\ref{fig:prompt_reviews_synth}) instructs the model to simulate a structured review grounded partially in the source text, while maintaining a natural and critical tone consistent with academic peer reviews. It emphasizes abstraction, editorial judgment, and partial grounding to avoid simple paraphrasing. Because the news articles from the WikinewsSum dataset were scraped directly from the webpage, they contain some artefacts that are not content of the article. Thus, we first clean that using GPT-4o-mini with the prompt template shown in Figure~\ref{fig:prompt_cleaning_news}. We chose GPT-4o-mini because it reliably removed social media links and scraping artifacts (based on manual checks), while being cost-effective. Next, we segment the cleaned articles using spaCy. For the choice of model used in synthetic document generation, we had three criteria: (1) strong general performance, (2) strong instruction-following (due to the specificity of our prompts), and (3) a long-context window (to handle long docs like papers). We selected DeepSeek-R1, as reasoning models showed strong prompt adherence. Importantly, we chose DeepSeek-R1 over models like OpenAI-o1 to avoid overlap with the OpenAI models used in the downstream classification task, thereby minimizing potential model-specific biases in evaluation \citep{liu-etal-2024-llms-narcissistic}. DeepSeek-R1 also offers comparable performance to OpenAI models on standard benchmarks \citep{DBLP:journals/corr/abs-2501-12948}. The generation prompt (Figure~\ref{fig:prompt_news_synth}) guides the model to produce synthetic news article covering the same topic, but differing in structure, tone, and emphasis to simulate diverse editorial styles. Both generation prompts include format specifications and grounding strategies. While these prompts may not be the most optimal, they produce sufficiently high-quality and controllable outputs for our specific goal of in the linking task.
Table~\ref{tab:synthetic_examples} provides illustrative examples of generated sentences and their corresponding linked natural sentences in both domains

\section{Human Evaluation Results}
\label{app:hum_eval_results}
\begin{figure*}[ht]
  \includegraphics[width=\linewidth]{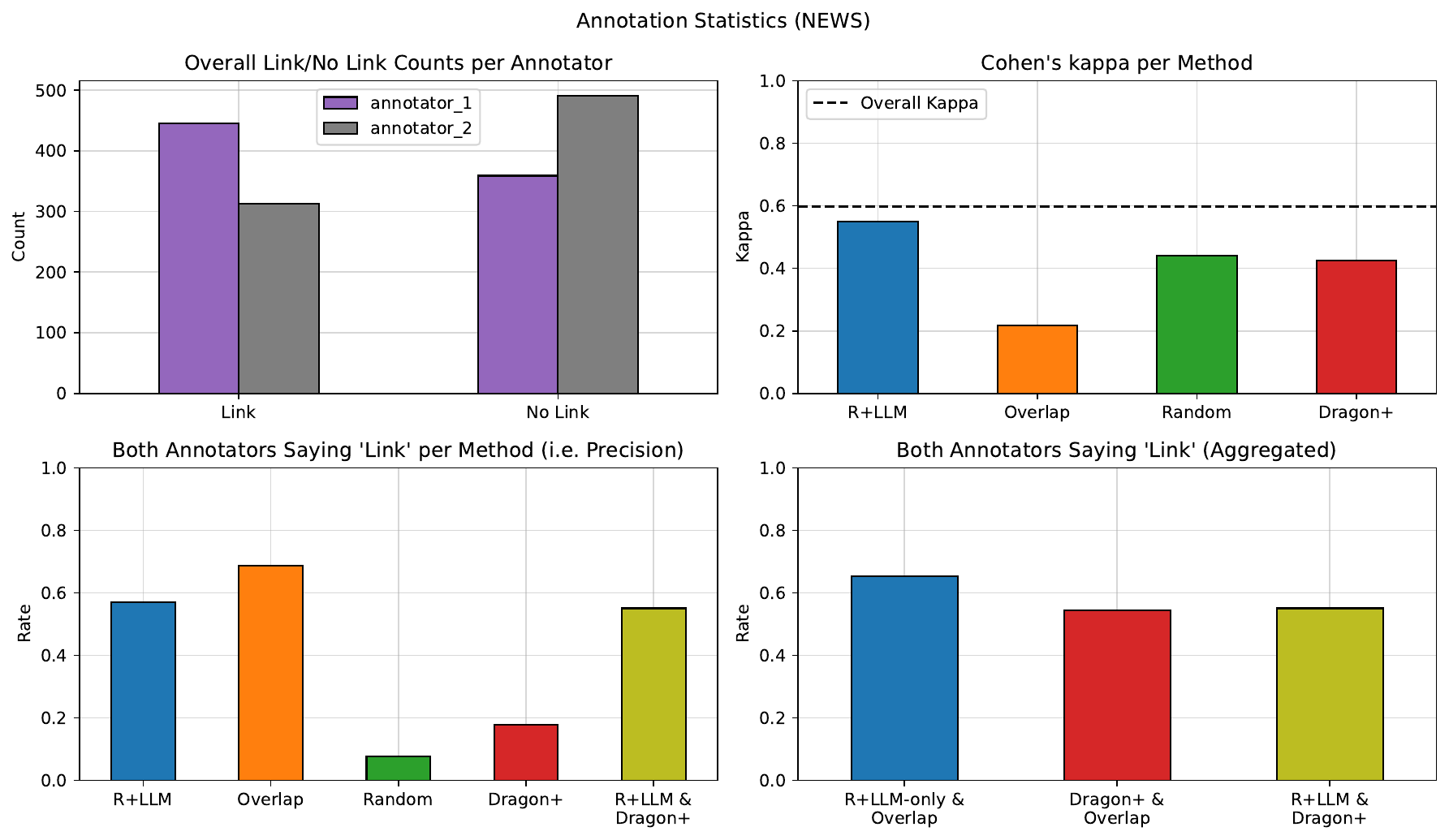}
  \caption {Annotation statistics in \texttt{NEWS-HE}.}
  \label{fig:annotation_statistics_news}
\end{figure*}

\begin{figure*}[ht]
  \includegraphics[width=\linewidth]{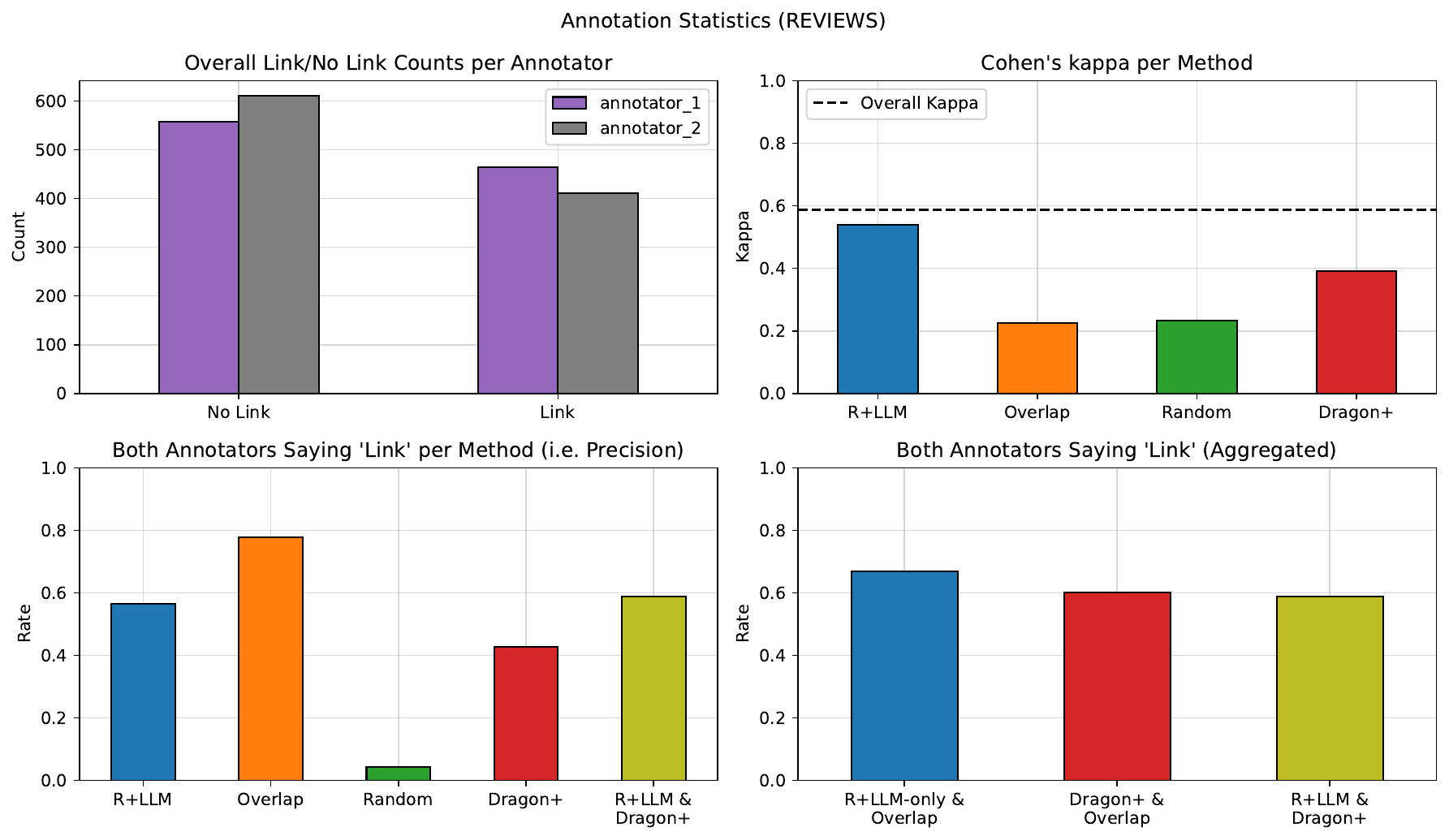}
  \caption {Annotation statistics in \texttt{REVIEWS-HE}.}
  \label{fig:annotation_statistics_reviews}
\end{figure*}

Figures \ref{fig:annotation_statistics_news} and \ref{fig:annotation_statistics_reviews} illustrate the results of the human evaluation in detail.

\section{Manual Recall Estimation Details}
\label{app:manual_recall}

To estimate recall, we sampled document pairs from \texttt{REVIEWS-HE} and \texttt{NEWS-HE}. For each domain, we selected random source sentences and exhaustively searched the target document for related sentences according to the domain-specific link definition. This was repeated until we labeled approximately 10\% of the total links: 102 in reviews and 80 in news. These links were used to compute precision, recall, and F1 for the top-performing \texttt{R+LLM}, retriever-only (\texttt{Dragon+}), the ``Both'' intersection strategy, and a random baseline. The manual annotation was done by one of the authors of this paper. 

The annotation of the 102 links in reviews took approx.\ 11 hours. The majority of the time was spent understanding the paper and the review, and ensuring that all possible target sentences were assessed. The 80 links in news took less time at approx.\ 5 hours due to the shorter documents, and the majority of the time was spent carefully reading the content of each sentence to judge whether it was related. This annotation further highlighted the sizable effort needed for full manual annotation of cross-document links, and the need for automatic methods to speed up and scale up the process.

\subsection{Alternative Evaluation under Candidate-Pool Constraints}

As noted in Section~\ref{sec:human_eval}, the main human evaluation (Table~\ref{tab:human_eval_link_agreement}) was conducted on top-$k$ candidate pools shown to annotators. To reconcile the difference between these acceptance rates and the exhaustive recall analysis in Table~\ref{tab:true_recall_results}, we repeated the recall estimation under the same top-$k$ constraints. The results are shown in Table~\ref{tab:true_recall_pool}.

\begin{table}[ht]
\centering
\small
\begin{tabular}{lccc}
\toprule
& \texttt{Recall} & \texttt{Precision} & \texttt{F1} \\
\midrule
\multicolumn{4}{l}{\textbf{\texttt{NEWS}}} \\
\midrule
\texttt{R+LLM}     & \texttt{0.77} & \texttt{0.93} & \texttt{0.82} \\
\texttt{Retriever} & \texttt{0.57} & \texttt{0.70} & \texttt{0.61} \\
\texttt{Both}      & \texttt{\textbf{0.68}} & \texttt{\textbf{0.95}} & \texttt{0.76} \\
\texttt{Random}    & \texttt{0.02} & \texttt{0.05} & \texttt{0.03} \\
\midrule
\multicolumn{4}{l}{\textbf{\texttt{REVIEWS}}} \\
\midrule
\texttt{R+LLM}     & \texttt{0.43} & \texttt{0.64} & \texttt{0.49} \\
\texttt{Retriever} & \texttt{0.18} & \texttt{0.28} & \texttt{0.21} \\
\texttt{Both}      & \texttt{0.24} & \texttt{0.52} & \texttt{0.32} \\
\texttt{Random}    & \texttt{0.00} & \texttt{0.00} & \texttt{0.00} \\
\bottomrule
\end{tabular}
\caption{Recall estimation under candidate-pool constraints, mimicking the human evaluation setup. ``Both'' regains the high precision observed in Table~\ref{tab:human_eval_link_agreement}, but at the cost of recall.}
\label{tab:true_recall_pool}
\end{table}

These results confirm that ``Both'' is best understood as a conservative high-precision filter: it produces links that are very likely to be correct, but its lower recall limits standalone utility. R+LLM remains the superior balanced approach when considering overall F1 under exhaustive conditions.

\subsection{Practical Guidelines for Practitioners}
Based on our experiments, our framework supports multiple operating modes depending on annotation budget and risk profile:

\begin{itemize}[noitemsep]
    \item \textbf{Single best method (R+LLM)}: Balanced precision/recall with moderate annotator load. Default choice when no strong constraints exist.
    \item \textbf{Intersection of top-N (``Both'')}: Conservative high-precision filter with reduced recall. Suitable when annotator time is very limited.
    \item \textbf{Union of top-N}: Maximizes recall at the cost of lower precision, producing larger candidate pools. Useful when coverage is critical and annotators can handle more candidates.
\end{itemize}

These modes can be selected to match application priorities, e.g., efficiency versus coverage.

\begin{figure}[t]
\centering
\begin{tcolorbox}[colback=gray!5, colframe=gray!80, fontupper=\ttfamily, title=News Cleaning Prompt, width=\linewidth]
\footnotesize
This is a scraped article from Wikinews. Due to the scraping, it may contain sentences that are image captions and social media links. Remove such sentences, but do not change the content of the article. Output the cleaned article in JSON format, where the key is 'cleaned\_article' and the value is the cleaned article text.\\
Input: \texttt{\{INPUT\}}
\end{tcolorbox}
    \caption{Prompt template for corrected scraped news articles.}
\label{fig:prompt_cleaning_news}
\end{figure}

\begin{figure*}[t]
\centering
\begin{tcolorbox}[colback=gray!5, colframe=gray!80, fontupper=\ttfamily, title=Synthetic Peer Review Generation Prompt, width=\linewidth]
\footnotesize
Task Overview\\
You are a peer reviewer evaluating a research paper in NLP. Your task is to write a realistic and well-rounded peer review for the paper.\\
Guidelines:\\
- Your review should be structured and natural, consisting of 8-12 sentences.\\
- 3 to 5 sentences should be implicitly grounded in specific ideas from the paper. These should express relevant critiques, observations, or praises without directly quoting or referencing the original text.\\
- The remaining sentences should address broader aspects such as clarity, methodology, contributions, generalization, writing quality, or suggestions for improvement.\\
- Do NOT explicitly cite, reference, or quote any sentence from the paper. The review should not be a direct commentary on specific lines.\\
How to Structure the Review:\\
- Rephrase, summarize, or abstract ideas: When addressing parts of the paper, reword and generalize instead of copying.\\
- Introduce new considerations: Some comments should reflect editorial judgment, unanswered questions, or high-level concerns rather than being tied to specific sentences.\\
- Omit some possible links: Not every review sentence should directly correspond to a sentence from the paper. Aim for a balanced mix of specific and general feedback.\\
- Rearrange information: The review’s flow should be different from the order of the paper to reflect a natural peer review process.\\
Input Format\\
You will receive a JSON object where:\\
- Each key is a sentence index from a paper section(s).\\
- Each value is the corresponding sentence text.\\
Output Format\\
- A peer review as a JSON object where:\\
\hspace*{2em}- Each key is a sentence index in the review (starting from 0).\\
\hspace*{2em}- Each value is the corresponding sentence text.\\
- A sentence mapping that links review sentences to the paper as a JSON object where:\\
\hspace*{2em}- Keys: Indices from the peer review.\\
\hspace*{2em}- Values: A list of corresponding indices from the paper section(s), or null if the sentence is not directly linked.\\
Example Input:\texttt{\{INPUT\_EXAMPLE\}}\\
Example Output:\texttt{\{OUTPUT\_EXAMPLE\}}\\
Input:\texttt{\{INPUT\}}\\
Output:
\end{tcolorbox}
    \caption{Prompt template for generating \texttt{REVIEWS-SYNTH}.}
\label{fig:prompt_reviews_synth}
\end{figure*}

\begin{figure*}[t]
\centering
\begin{tcolorbox}[colback=gray!5, colframe=gray!80, fontupper=\ttfamily, title=Synthetic News Generation Prompt, width=\linewidth]
\footnotesize
Task Overview:\\
You are generating a news article that covers the same event or topic as an original article presenting it from a different editorial angle. The goal is to simulate how separate news organizations might independently report on the same subject, differing in structure, tone, detail, and emphasis. The new article should be a plausible alternative version of coverage on the same topic, not a direct rephrasing or summary of the original.\\
Guidelines:\\
- The article should be realistic, coherent, and reflective of a distinctive voice or editorial style.\\
- 3 to 5 sentences in the new article should reflect content from the original. These sentences may describe similar facts, events, or issues, but using different wording, tone, or framing.\\
- Do not replicate the original article's sentence-by-sentence structure or closely paraphrase its content.\\
- The remaining sentences should introduce: New but plausible perspectives, context, or editorial framing. Additional background or expert input. A different narrative structure or omission of certain original points.\\
- The new article must have a different length from the original, but not be much shorter than the original.\\
Variation Strategies:\\
- Rephrase and shift style: Change vocabulary, sentence structure, or writing tone to reflect a different editorial voice.\\
- Frame the topic differently: Adjust emphasis or viewpoint—, for example, highlighting controversy, local impact, or long-term implications.\\
- Add or omit information: Introduce plausible context, background, or expert input, or skip less relevant details from the original.\\
- Reorganize the narrative: Present the information in a different order to create a new logical or rhetorical flow.\\
Input Format:\\
You will receive a JSON object where:\\
- Each key is a sentence index from the original article.\\
- Each value is the corresponding sentence text.\\
Output Format:\\
- The generated news articlea as a JSON object:\\
    - Keys: Indices of sentences in the new article (starting from 0).\\
    - Values: The text of each sentence.\\
A sentence mapping that links sentences in the new article to sentences in the original article as a JSON object where:\\
    - Keys: Indices of sentences in the new article.\\
    - Values: A list of sentence indices from the original that the new sentence relates to, or null if it is not directly linked to any original sentence.\\
Example input: \texttt{\{INPUT\_EXAMPLE\}}\\
Example Output: \texttt{\{OUTPUT\_EXAMPLE\}}\\
Input: \texttt{\{INPUT\}}\\
Output:
\end{tcolorbox}
    \caption{Prompt template for generating \texttt{NEWS-SYNTH}.}
\label{fig:prompt_news_synth}
\end{figure*}

\begin{table*}[ht]
\centering
\resizebox{\textwidth}{!}{
\begin{tabular}{@{}lccc!{\color{gray}\vrule width 0.6pt}ccc!{\color{gray}\vrule width 0.6pt}ccc!{\color{gray}\vrule width 0.6pt}ccc!{\color{gray}\vrule width 0.6pt}ccc!{\color{gray}\vrule width 0.6pt}ccc@{}}
\toprule
\texttt{Model} & \multicolumn{3}{c}{\texttt{K=1}} & \multicolumn{3}{c}{\texttt{K=3}} & \multicolumn{3}{c}{\texttt{K=5}} & \multicolumn{3}{c}{\texttt{K=7}} & \multicolumn{3}{c}{\texttt{K=10}} & \multicolumn{3}{c}{\texttt{K=20}} \\
 & \texttt{P} & \texttt{R} & \texttt{F1} & \texttt{P} & \texttt{R} & \texttt{F1} & \texttt{P} & \texttt{R} & \texttt{F1} & \texttt{P} & \texttt{R} & \texttt{F1} & \texttt{P} & \texttt{R} & \texttt{F1} & \texttt{P} & \texttt{R} & \texttt{F1} \\
\midrule
\texttt{bm25} & \texttt{56.71} & \texttt{34.12} & \texttt{40.91} & \texttt{37.21} & \texttt{63.17} & \texttt{45.03} & \texttt{28.75} & \texttt{76.74} & \texttt{40.28} & \texttt{24.17} & \texttt{83.59} & \texttt{35.95} & \texttt{21.17} & \texttt{89.40} & \texttt{32.43} & \texttt{18.72} & \texttt{96.40} & \texttt{28.95} \\
\texttt{splade} & \texttt{62.03} & \texttt{37.47} & \texttt{44.87} & \texttt{41.60} & \texttt{70.11} & \texttt{50.25} & \texttt{30.74} & \texttt{81.93} & \texttt{43.09} & \texttt{25.44} & \texttt{88.37} & \texttt{37.92} & \texttt{21.88} & \texttt{93.23} & \texttt{33.60} & \texttt{18.90} & \texttt{98.26} & \texttt{29.27} \\
\texttt{bgem3-sparse} & \texttt{59.20} & \texttt{35.30} & \texttt{42.47} & \texttt{39.94} & \texttt{67.59} & \texttt{48.30} & \texttt{30.07} & \texttt{80.30} & \texttt{42.17} & \texttt{25.07} & \texttt{86.97} & \texttt{37.35} & \texttt{21.70} & \texttt{92.21} & \texttt{33.31} & \texttt{18.88} & \texttt{98.07} & \texttt{29.24} \\
\midrule
\texttt{sfr} & \texttt{69.14} & \texttt{41.03} & \texttt{49.50} & \texttt{48.34} & \texttt{80.72} & \texttt{58.19} & \texttt{34.51} & \texttt{91.97} & \texttt{48.38} & \texttt{27.49} & \texttt{95.85} & \texttt{41.04} & \texttt{22.81} & \texttt{98.01} & \texttt{35.13} & \texttt{19.02} & \texttt{99.68} & \texttt{29.50} \\
\texttt{all-mpnet} & \texttt{65.08} & \texttt{39.40} & \texttt{47.12} & \texttt{45.81} & \texttt{77.36} & \texttt{55.35} & \texttt{33.23} & \texttt{88.90} & \texttt{46.62} & \texttt{26.76} & \texttt{93.52} & \texttt{39.95} & \texttt{22.48} & \texttt{96.52} & \texttt{34.60} & \texttt{18.99} & \texttt{99.36} & \texttt{29.44} \\
\texttt{bgem3-dense} & \texttt{67.65} & \texttt{41.05} & \texttt{49.06} & \texttt{45.55} & \texttt{76.86} & \texttt{55.01} & \texttt{33.03} & \texttt{88.13} & \texttt{46.32} & \texttt{26.59} & \texttt{92.78} & \texttt{39.69} & \texttt{22.44} & \texttt{96.23} & \texttt{34.53} & \texttt{18.95} & \texttt{98.95} & \texttt{29.38} \\
\texttt{contriever} & \texttt{64.13} & \texttt{38.85} & \texttt{46.46} & \texttt{43.38} & \texttt{73.45} & \texttt{52.46} & \texttt{32.16} & \texttt{85.82} & \texttt{45.10} & \texttt{26.26} & \texttt{91.65} & \texttt{39.18} & \texttt{22.35} & \texttt{95.73} & \texttt{34.38} & \texttt{18.99} & \texttt{99.34} & \texttt{29.45} \\
\texttt{dragon\_plus} & \texttt{66.67} & \texttt{40.31} & \texttt{48.21} & \texttt{45.82} & \texttt{77.36} & \texttt{55.36} & \texttt{33.27} & \texttt{89.02} & \texttt{46.68} & \texttt{26.86} & \texttt{93.87} & \texttt{40.10} & \texttt{22.53} & \texttt{96.72} & \texttt{34.67} & \texttt{19.01} & \texttt{99.50} & \texttt{29.48} \\
\midrule
\texttt{bge-reranker} & \texttt{61.29} & \texttt{36.29} & \texttt{43.78} & \texttt{44.86} & \texttt{75.38} & \texttt{54.10} & \texttt{33.03} & \texttt{88.05} & \texttt{46.29} & \texttt{26.78} & \texttt{93.33} & \texttt{39.96} & \texttt{22.56} & \texttt{96.74} & \texttt{34.72} & \texttt{19.02} & \texttt{99.69} & \texttt{29.50} \\
\texttt{ms\_marco\_minilm} & \texttt{65.06} & \texttt{39.03} & \texttt{46.83} & \texttt{45.24} & \texttt{76.24} & \texttt{54.63} & \texttt{32.84} & \texttt{87.76} & \texttt{46.07} & \texttt{26.66} & \texttt{93.07} & \texttt{39.80} & \texttt{22.45} & \texttt{96.27} & \texttt{34.55} & \texttt{18.98} & \texttt{99.25} & \texttt{29.43} \\
\bottomrule
\end{tabular}}
\caption{Results on \texttt{NEWS-ECB+}. Precision (P), Recall (R), and F1 at different cutoffs (K).}
\label{tab:news_ecb}
\end{table*}

\begin{table*}[ht]
\centering
\resizebox{\textwidth}{!}{
\begin{tabular}{@{}lccc!{\color{gray}\vrule width 0.6pt}ccc!{\color{gray}\vrule width 0.6pt}ccc!{\color{gray}\vrule width 0.6pt}ccc!{\color{gray}\vrule width 0.6pt}ccc!{\color{gray}\vrule width 0.6pt}ccc@{}}
\toprule
\texttt{Model} & \multicolumn{3}{c}{\texttt{K=1}} & \multicolumn{3}{c}{\texttt{K=3}} & \multicolumn{3}{c}{\texttt{K=5}} & \multicolumn{3}{c}{\texttt{K=7}} & \multicolumn{3}{c}{\texttt{K=10}} & \multicolumn{3}{c}{\texttt{K=20}} \\
 & \texttt{P} & \texttt{R} & \texttt{F1} & \texttt{P} & \texttt{R} & \texttt{F1} & \texttt{P} & \texttt{R} & \texttt{F1} & \texttt{P} & \texttt{R} & \texttt{F1} & \texttt{P} & \texttt{R} & \texttt{F1} & \texttt{P} & \texttt{R} & \texttt{F1} \\
\midrule
\texttt{bm25} & \texttt{65.29} & \texttt{60.26} & \texttt{61.78} & \texttt{29.75} & \texttt{78.60} & \texttt{42.25} & \texttt{19.75} & \texttt{86.12} & \texttt{31.52} & \texttt{14.64} & \texttt{89.23} & \texttt{24.73} & \texttt{11.14} & \texttt{94.35} & \texttt{19.63} & \texttt{8.63} & \texttt{99.59} & \texttt{15.56} \\
\texttt{splade} & \texttt{67.36} & \texttt{61.57} & \texttt{63.36} & \texttt{29.48} & \texttt{78.39} & \texttt{42.00} & \texttt{20.17} & \texttt{87.16} & \texttt{32.08} & \texttt{15.05} & \texttt{91.43} & \texttt{25.41} & \texttt{11.26} & \texttt{95.45} & \texttt{19.85} & \texttt{8.61} & \texttt{99.17} & \texttt{15.52} \\
\texttt{bgem3-sparse} & \texttt{71.49} & \texttt{65.85} & \texttt{67.56} & \texttt{31.96} & \texttt{84.86} & \texttt{45.50} & \texttt{20.41} & \texttt{88.95} & \texttt{32.56} & \texttt{15.35} & \texttt{93.03} & \texttt{25.89} & \texttt{11.39} & \texttt{96.49} & \texttt{20.07} & \texttt{8.63} & \texttt{99.59} & \texttt{15.56} \\
\midrule
\texttt{sfr} & \texttt{74.79} & \texttt{67.99} & \texttt{70.04} & \texttt{34.71} & \texttt{90.11} & \texttt{48.91} & \texttt{22.31} & \texttt{95.23} & \texttt{35.35} & \texttt{16.29} & \texttt{97.18} & \texttt{27.37} & \texttt{11.59} & \texttt{97.80} & \texttt{20.42} & \texttt{8.65} & \texttt{100.00} & \texttt{15.60} \\
\texttt{all-mpnet} & \texttt{76.86} & \texttt{70.47} & \texttt{72.38} & \texttt{34.57} & \texttt{90.79} & \texttt{48.99} & \texttt{22.15} & \texttt{96.17} & \texttt{35.30} & \texttt{16.35} & \texttt{97.85} & \texttt{27.50} & \texttt{11.72} & \texttt{98.97} & \texttt{20.64} & \texttt{8.65} & \texttt{100.00} & \texttt{15.60} \\
\texttt{bgem3-dense} & \texttt{79.34} & \texttt{72.53} & \texttt{74.59} & \texttt{35.26} & \texttt{92.11} & \texttt{49.83} & \texttt{22.23} & \texttt{95.43} & \texttt{35.30} & \texttt{16.53} & \texttt{99.04} & \texttt{27.80} & \texttt{11.80} & \texttt{99.79} & \texttt{20.79} & \texttt{8.65} & \texttt{100.00} & \texttt{15.60} \\
\texttt{contriever} & \texttt{65.70} & \texttt{60.48} & \texttt{62.05} & \texttt{30.58} & \texttt{81.16} & \texttt{43.51} & \texttt{19.92} & \texttt{86.68} & \texttt{31.76} & \texttt{15.35} & \texttt{92.27} & \texttt{25.84} & \texttt{11.51} & \texttt{97.31} & \texttt{20.27} & \texttt{8.65} & \texttt{100.00} & \texttt{15.60} \\
\texttt{dragon\_plus} & \texttt{78.51} & \texttt{72.52} & \texttt{74.38} & \texttt{34.71} & \texttt{91.54} & \texttt{49.29} & \texttt{21.57} & \texttt{93.90} & \texttt{34.42} & \texttt{16.41} & \texttt{98.47} & \texttt{27.61} & \texttt{11.72} & \texttt{99.17} & \texttt{20.64} & \texttt{8.65} & \texttt{100.00} & \texttt{15.60} \\
\midrule
\texttt{bge-reranker} & \texttt{76.03} & \texttt{70.12} & \texttt{71.90} & \texttt{34.85} & \texttt{90.66} & \texttt{49.17} & \texttt{22.40} & \texttt{95.85} & \texttt{35.51} & \texttt{16.65} & \texttt{99.17} & \texttt{27.96} & \texttt{11.84} & \texttt{100.00} & \texttt{20.86} & \texttt{8.65} & \texttt{100.00} & \texttt{15.60} \\
\texttt{ms\_marco\_minilm} & \texttt{82.64} & \texttt{75.56} & \texttt{77.69} & \texttt{33.88} & \texttt{88.79} & \texttt{47.98} & \texttt{21.90} & \texttt{94.46} & \texttt{34.83} & \texttt{16.12} & \texttt{96.75} & \texttt{27.13} & \texttt{11.72} & \texttt{98.76} & \texttt{20.63} & \texttt{8.65} & \texttt{100.00} & \texttt{15.60} \\
\bottomrule
\end{tabular}}
\caption{Results on \texttt{NEWS-SYNTH}. Precision (P), Recall (R), and F1 at different cutoffs (K).}
\label{tab:news_synth}
\end{table*}

\begin{table*}[ht]
\centering
\resizebox{\textwidth}{!}{
\begin{tabular}{@{}lccc!{\color{gray}\vrule width 0.6pt}ccc!{\color{gray}\vrule width 0.6pt}ccc!{\color{gray}\vrule width 0.6pt}ccc!{\color{gray}\vrule width 0.6pt}ccc!{\color{gray}\vrule width 0.6pt}ccc@{}}
\toprule
\texttt{Model} & \multicolumn{3}{c}{\texttt{K=1}} & \multicolumn{3}{c}{\texttt{K=3}} & \multicolumn{3}{c}{\texttt{K=5}} & \multicolumn{3}{c}{\texttt{K=7}} & \multicolumn{3}{c}{\texttt{K=10}} & \multicolumn{3}{c}{\texttt{K=20}} \\
 & \texttt{P} & \texttt{R} & \texttt{F1} & \texttt{P} & \texttt{R} & \texttt{F1} & \texttt{P} & \texttt{R} & \texttt{F1} & \texttt{P} & \texttt{R} & \texttt{F1} & \texttt{P} & \texttt{R} & \texttt{F1} & \texttt{P} & \texttt{R} & \texttt{F1} \\
\midrule
\texttt{bm25} & \texttt{28.63} & \texttt{16.70} & \texttt{19.93} & \texttt{19.24} & \texttt{30.57} & \texttt{22.21} & \texttt{16.12} & \texttt{41.64} & \texttt{22.03} & \texttt{13.09} & \texttt{47.14} & \texttt{19.57} & \texttt{10.97} & \texttt{55.01} & \texttt{17.64} & \texttt{6.83} & \texttt{66.39} & \texttt{12.11} \\
\texttt{splade} & \texttt{25.11} & \texttt{14.49} & \texttt{17.27} & \texttt{20.85} & \texttt{33.58} & \texttt{24.16} & \texttt{16.56} & \texttt{42.65} & \texttt{22.66} & \texttt{14.10} & \texttt{51.12} & \texttt{21.12} & \texttt{11.72} & \texttt{60.04} & \texttt{18.92} & \texttt{7.14} & \texttt{69.83} & \texttt{12.65} \\
\texttt{bgem3-sparse} & \texttt{26.87} & \texttt{14.79} & \texttt{17.93} & \texttt{22.17} & \texttt{36.03} & \texttt{25.79} & \texttt{16.74} & \texttt{44.47} & \texttt{23.09} & \texttt{13.91} & \texttt{49.90} & \texttt{20.79} & \texttt{11.32} & \texttt{56.08} & \texttt{18.15} & \texttt{7.11} & \texttt{70.08} & \texttt{12.62} \\
\midrule
\texttt{sfr} & \texttt{24.67} & \texttt{13.19} & \texttt{16.20} & \texttt{20.12} & \texttt{31.34} & \texttt{23.09} & \texttt{16.56} & \texttt{43.01} & \texttt{22.64} & \texttt{13.97} & \texttt{49.70} & \texttt{20.81} & \texttt{11.59} & \texttt{58.52} & \texttt{18.64} & \texttt{7.80} & \texttt{75.93} & \texttt{13.82} \\
\texttt{all-mpnet} & \texttt{27.75} & \texttt{15.10} & \texttt{18.40} & \texttt{17.77} & \texttt{27.94} & \texttt{20.31} & \texttt{15.07} & \texttt{38.93} & \texttt{20.58} & \texttt{12.96} & \texttt{46.37} & \texttt{19.36} & \texttt{10.88} & \texttt{54.90} & \texttt{17.52} & \texttt{6.96} & \texttt{68.69} & \texttt{12.36} \\
\texttt{bgem3-dense} & \texttt{29.96} & \texttt{16.56} & \texttt{20.15} & \texttt{22.32} & \texttt{35.17} & \texttt{25.73} & \texttt{18.68} & \texttt{47.51} & \texttt{25.48} & \texttt{15.61} & \texttt{55.51} & \texttt{23.31} & \texttt{12.78} & \texttt{64.00} & \texttt{20.53} & \texttt{7.84} & \texttt{77.14} & \texttt{13.91} \\
\texttt{contriever} & \texttt{24.67} & \texttt{13.27} & \texttt{16.34} & \texttt{18.21} & \texttt{28.16} & \texttt{20.88} & \texttt{14.10} & \texttt{36.51} & \texttt{19.33} & \texttt{12.59} & \texttt{45.04} & \texttt{18.86} & \texttt{10.18} & \texttt{51.32} & \texttt{16.43} & \texttt{6.74} & \texttt{66.69} & \texttt{11.98} \\
\texttt{dragon\_plus} & \texttt{30.84} & \texttt{17.78} & \texttt{21.28} & \texttt{21.73} & \texttt{36.03} & \texttt{25.48} & \texttt{18.33} & \texttt{48.69} & \texttt{25.24} & \texttt{15.10} & \texttt{55.14} & \texttt{22.65} & \texttt{11.63} & \texttt{59.28} & \texttt{18.74} & \texttt{7.49} & \texttt{73.50} & \texttt{13.29} \\
\midrule
\texttt{bge-reranker} & \texttt{23.35} & \texttt{12.34} & \texttt{15.18} & \texttt{20.41} & \texttt{34.16} & \texttt{23.98} & \texttt{16.12} & \texttt{43.83} & \texttt{22.33} & \texttt{14.73} & \texttt{53.11} & \texttt{22.06} & \texttt{12.16} & \texttt{60.70} & \texttt{19.54} & \texttt{7.91} & \texttt{77.11} & \texttt{14.02} \\
\texttt{ms\_marco\_minilm} & \texttt{33.92} & \texttt{20.31} & \texttt{23.96} & \texttt{23.64} & \texttt{38.66} & \texttt{27.60} & \texttt{18.59} & \texttt{48.73} & \texttt{25.52} & \texttt{15.10} & \texttt{53.58} & \texttt{22.53} & \texttt{11.94} & \texttt{59.51} & \texttt{19.19} & \texttt{7.42} & \texttt{72.97} & \texttt{13.17} \\
\bottomrule
\end{tabular}}
\caption{Results on \texttt{REVIEWS-SYNTH}. Precision (P), Recall (R), and F1 at different cutoffs (K).}
\label{tab:reviews_synth}
\end{table*}

\begin{table*}[ht]
\centering
\resizebox{\textwidth}{!}{
\begin{tabular}{@{}lccc!{\color{gray}\vrule width 0.6pt}ccc!{\color{gray}\vrule width 0.6pt}ccc!{\color{gray}\vrule width 0.6pt}ccc!{\color{gray}\vrule width 0.6pt}ccc!{\color{gray}\vrule width 0.6pt}ccc@{}}
\toprule
\texttt{Model} & \multicolumn{3}{c}{\texttt{K=1}} & \multicolumn{3}{c}{\texttt{K=3}} & \multicolumn{3}{c}{\texttt{K=5}} & \multicolumn{3}{c}{\texttt{K=7}} & \multicolumn{3}{c}{\texttt{K=10}} & \multicolumn{3}{c}{\texttt{K=20}} \\
 & \texttt{P} & \texttt{R} & \texttt{F1} & \texttt{P} & \texttt{R} & \texttt{F1} & \texttt{P} & \texttt{R} & \texttt{F1} & \texttt{P} & \texttt{R} & \texttt{F1} & \texttt{P} & \texttt{R} & \texttt{F1} & \texttt{P} & \texttt{R} & \texttt{F1} \\
\midrule
\texttt{bm25} & \texttt{58.28} & \texttt{53.15} & \texttt{54.75} & \texttt{28.21} & \texttt{73.84} & \texttt{40.00} & \texttt{18.75} & \texttt{81.30} & \texttt{29.95} & \texttt{14.15} & \texttt{85.36} & \texttt{23.91} & \texttt{10.29} & \texttt{88.51} & \texttt{18.22} & \texttt{5.51} & \texttt{93.90} & \texttt{10.33} \\
\texttt{splade} & \texttt{54.36} & \texttt{49.33} & \texttt{50.91} & \texttt{28.31} & \texttt{73.68} & \texttt{40.06} & \texttt{18.94} & \texttt{81.04} & \texttt{30.15} & \texttt{14.30} & \texttt{85.51} & \texttt{24.12} & \texttt{10.66} & \texttt{90.47} & \texttt{18.82} & \texttt{5.64} & \texttt{95.59} & \texttt{10.57} \\
\texttt{bgem3-sparse} & \texttt{57.49} & \texttt{52.53} & \texttt{54.11} & \texttt{28.57} & \texttt{74.61} & \texttt{40.49} & \texttt{18.94} & \texttt{81.68} & \texttt{30.23} & \texttt{14.31} & \texttt{85.99} & \texttt{24.17} & \texttt{10.49} & \texttt{89.20} & \texttt{18.53} & \texttt{5.62} & \texttt{95.38} & \texttt{10.54} \\
\midrule
\texttt{sfr} & \texttt{45.94} & \texttt{40.83} & \texttt{42.44} & \texttt{24.39} & \texttt{62.65} & \texttt{34.34} & \texttt{16.96} & \texttt{72.01} & \texttt{26.94} & \texttt{13.08} & \texttt{77.95} & \texttt{22.05} & \texttt{9.82} & \texttt{82.82} & \texttt{17.34} & \texttt{5.46} & \texttt{92.05} & \texttt{10.23} \\
\texttt{all-mpnet} & \texttt{43.98} & \texttt{39.56} & \texttt{40.94} & \texttt{23.21} & \texttt{60.00} & \texttt{32.76} & \texttt{16.16} & \texttt{69.13} & \texttt{25.71} & \texttt{12.59} & \texttt{75.13} & \texttt{21.23} & \texttt{9.46} & \texttt{80.34} & \texttt{16.71} & \texttt{5.26} & \texttt{89.16} & \texttt{9.86} \\
\texttt{bgem3-dense} & \texttt{57.30} & \texttt{51.52} & \texttt{53.33} & \texttt{28.80} & \texttt{74.30} & \texttt{40.62} & \texttt{19.28} & \texttt{82.02} & \texttt{30.62} & \texttt{14.64} & \texttt{87.07} & \texttt{24.66} & \texttt{10.72} & \texttt{91.02} & \texttt{18.94} & \texttt{5.67} & \texttt{96.14} & \texttt{10.63} \\
\texttt{contriever} & \texttt{46.91} & \texttt{42.36} & \texttt{43.79} & \texttt{24.49} & \texttt{63.80} & \texttt{34.68} & \texttt{17.43} & \texttt{74.47} & \texttt{27.73} & \texttt{13.40} & \texttt{79.65} & \texttt{22.58} & \texttt{10.06} & \texttt{85.49} & \texttt{17.77} & \texttt{5.50} & \texttt{93.45} & \texttt{10.31} \\
\texttt{dragon\_plus} & \texttt{63.08} & \texttt{56.75} & \texttt{58.73} & \texttt{30.33} & \texttt{77.79} & \texttt{42.67} & \texttt{20.20} & \texttt{86.17} & \texttt{32.11} & \texttt{15.10} & \texttt{89.80} & \texttt{25.44} & \texttt{10.95} & \texttt{93.02} & \texttt{19.35} & \texttt{5.73} & \texttt{97.08} & \texttt{10.74} \\
\midrule
\texttt{bge-reranker} & \texttt{47.50} & \texttt{43.42} & \texttt{44.71} & \texttt{26.15} & \texttt{68.26} & \texttt{37.06} & \texttt{18.20} & \texttt{78.30} & \texttt{29.02} & \texttt{13.92} & \texttt{83.32} & \texttt{23.50} & \texttt{10.43} & \texttt{88.60} & \texttt{18.43} & \texttt{5.65} & \texttt{95.74} & \texttt{10.58} \\
\texttt{ms\_marco\_minilm} & \texttt{64.94} & \texttt{59.24} & \texttt{61.03} & \texttt{30.43} & \texttt{79.37} & \texttt{43.09} & \texttt{20.12} & \texttt{86.37} & \texttt{32.04} & \texttt{15.00} & \texttt{89.68} & \texttt{25.30} & \texttt{10.84} & \texttt{92.41} & \texttt{19.16} & \texttt{5.68} & \texttt{96.65} & \texttt{10.65} \\
\bottomrule
\end{tabular}}
\caption{Results on \texttt{REVIEWS-F1000}. Precision (P), Recall (R), and F1 at different cutoffs (K).}
\label{tab:reviews_F1000}
\end{table*}

\begin{table*}[!t]
\centering
\resizebox{\textwidth}{!}{
\begin{tabular}{p{0.14\textwidth} p{0.42\textwidth} p{0.42\textwidth}}
\toprule
\textbf{Domain} & \textbf{Source Sentence (Generated)} & \textbf{Target Sentence (Natural)} \\
\midrule
\texttt{NEWS} &
In Southern Europe, the ripple effects of the disrupted flight schedules are being felt widely, with airlines cautioning that recovery from this crisis could take days or even weeks. &
Some flights from Spain and Portugal, together with upwards of 4,000 flights across Northern Europe, have been affected, and the knock-on effect of aircraft and crews out of position could disrupt air travel worldwide for up to 72 hours. \\
\midrule
\texttt{NEWS} &
Newsweek’s Steven Levy has pointed out that top-ranked blogs remain overwhelmingly white and male, underscoring a digital divide as real as any glass ceiling. &
Steven Levy, a senior editor of Newsweek, has recently written a column addressing concerns about the over-representation of white males among top bloggers on the Internet. \\
\midrule
\texttt{PEER-REVIEWS} &
The inclusion of negative samples in the supervised fine-tuning pipeline appears to be non-trivial, as this variant underperforms relative to other supervised strategies. &
Among all supervised methods, the SFT with negatives performs the worst, showing that using negative feedback in supervised training analogically to preference optimization is non-trivial. \\
\midrule
\texttt{PEER-REVIEWS} &
Coverage limitations of some rule-based components mean that a sizable fraction of utterances remain unhandled, which may impact the overall robustness. &
P1.1 and P2 have to meet certain conditions to be applied, therefore they do not have full coverage of CS data: 36\% and 31\% for SEAME, 60\% and 58\% for Miami. \\
\bottomrule
\end{tabular}}
\caption{Examples of sentence-level links in the \texttt{NEWS-SYNTH} and \texttt{REVIEWS-SYNTH} datasets. Source sentences are generated using an LLM conditioned on the target document, with sentence-level links specified during generation.}
\label{tab:synthetic_examples}
\end{table*}

\end{document}